\documentclass[11pt]{article}

\usepackage[final]{acl}

\usepackage{times}
\usepackage{latexsym}

\usepackage[T1]{fontenc}

\usepackage[utf8]{inputenc}

\usepackage{microtype}

\usepackage{inconsolata}

\usepackage{graphicx}
\usepackage{amsmath,amssymb}
\usepackage[normalem]{ulem}   

\usepackage{booktabs}
\usepackage{multirow}
\usepackage{colortbl}
\usepackage{xcolor}
\usepackage{tabularx}

\usepackage{tcolorbox}
\tcbuselibrary{skins, breakable} 
\usepackage{enumitem}

 \usepackage{makecell}
\usepackage{pifont}

\usepackage{svg}
\usepackage{url}

\makeatletter
\newcommand\blfootnote[1]{%
  \begingroup
    \renewcommand\thefootnote{}%
    \renewcommand\@makefnmark{}%
    \footnotetext{#1}%
  \endgroup
}
\makeatother

\title{Don't Guess, Ask Wisely: Benchmarking and Training Proactive Elicitation in Open-Domain LLM Conversations}

\author{
  \textbf{Rongyu Zhang}\textsuperscript{1*},
  \textbf{Mutian Bao}\textsuperscript{1*},
  \textbf{Jinjian Zhang}\textsuperscript{2*},
  \textbf{Yuetian Du}\textsuperscript{1},
  \textbf{He Xu}\textsuperscript{1},
  \textbf{Wei Zhou}\textsuperscript{2},
\\
  \textbf{Xiaojun Zhu}\textsuperscript{1},
  \textbf{Ming Kong}\textsuperscript{1},
  \textbf{Jie Liu}\textsuperscript{3$\dagger$},
  \textbf{Linjian Mo}\textsuperscript{2$\dagger$},
  \textbf{Qiang Zhu}\textsuperscript{1$\dagger$}
\\
\\
  \textsuperscript{1}Zhejiang University \quad
  \textsuperscript{2}Ant Group \quad
  \textsuperscript{3}City University of Hong Kong
\\
  \texttt{\{rongyu.zhang, zhuq\}@zju.edu.cn}
\\[0.6em]
  {\small
  \href{https://github.com/zhangrongyu0101/VagueBench}{\raisebox{-0.18em}{\includegraphics[height=1.25em]{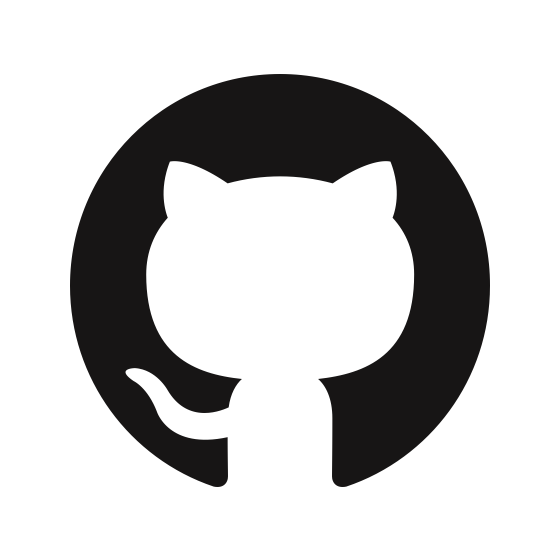}}\,\textbf{Code}}
  \qquad
  \href{https://huggingface.co/datasets/rongyu0101/VagueBench}{\raisebox{-0.22em}{\includegraphics[height=1.55em]{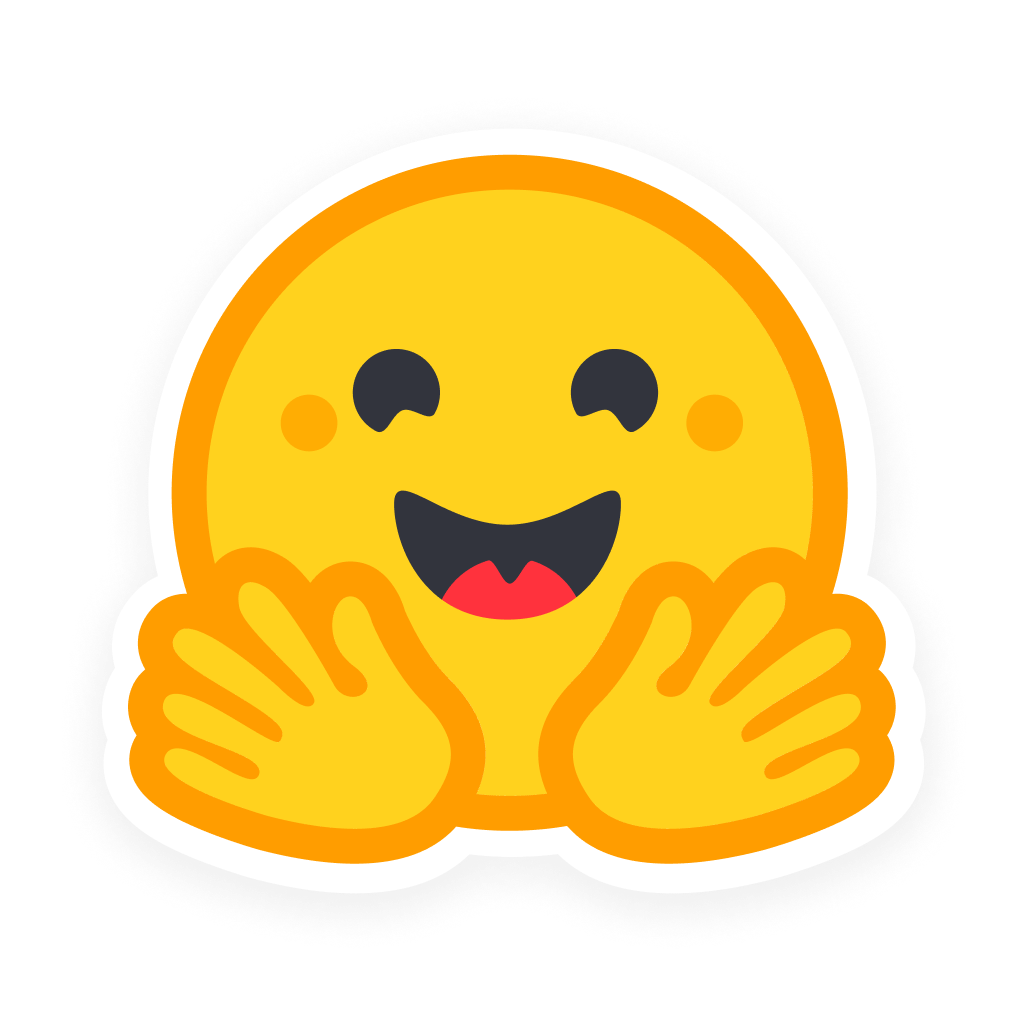}}\,\textbf{Data}}}
}

\begin{document}
\maketitle

\blfootnote{%
  \begin{tabular}{@{}l@{}}
  \llap{$^{*}$}Equal contribution.\\
  \llap{$^{\dagger}$}Corresponding authors.
  \end{tabular}%
}


\begin{abstract}
When user queries are incomplete, large language models should ask targeted questions before committing to an answer, a capability we call proactive elicitation, yet evaluation on real user-assistant queries across seven LLMs (8B to frontier) shows that fewer than 11\% of underspecified queries trigger a follow-up question. To benchmark and train this capability, we introduce VagueBench, where each seed query is annotated with structured intent and detail units then degraded into intent-vague and fully-vague variants via controlled removal of linguistic cues, yielding known hidden information sets for evaluating both when and what to ask. We propose DialogGRPO, an offline reinforcement learning method that organizes multi-turn elicitation as a dialogue-state directed acyclic graph, where candidate questions at each node receive dense turn-level rewards combining immediate information gain with lookahead value through group-relative comparison, enabling fine-grained credit assignment without runtime simulation. DialogGRPO elicits missing information on over 88\% of vague queries with substantial gains in both recovery efficiency and answer coverage, without over-asking on complete queries, outperforms prompting and test-time-scaling baselines, and generalizes zero-shot to MultiWOZ 2.4 and AmbigQA.
\end{abstract}


\begin{figure}[t!]
    \centering
    \includegraphics[width=\columnwidth]{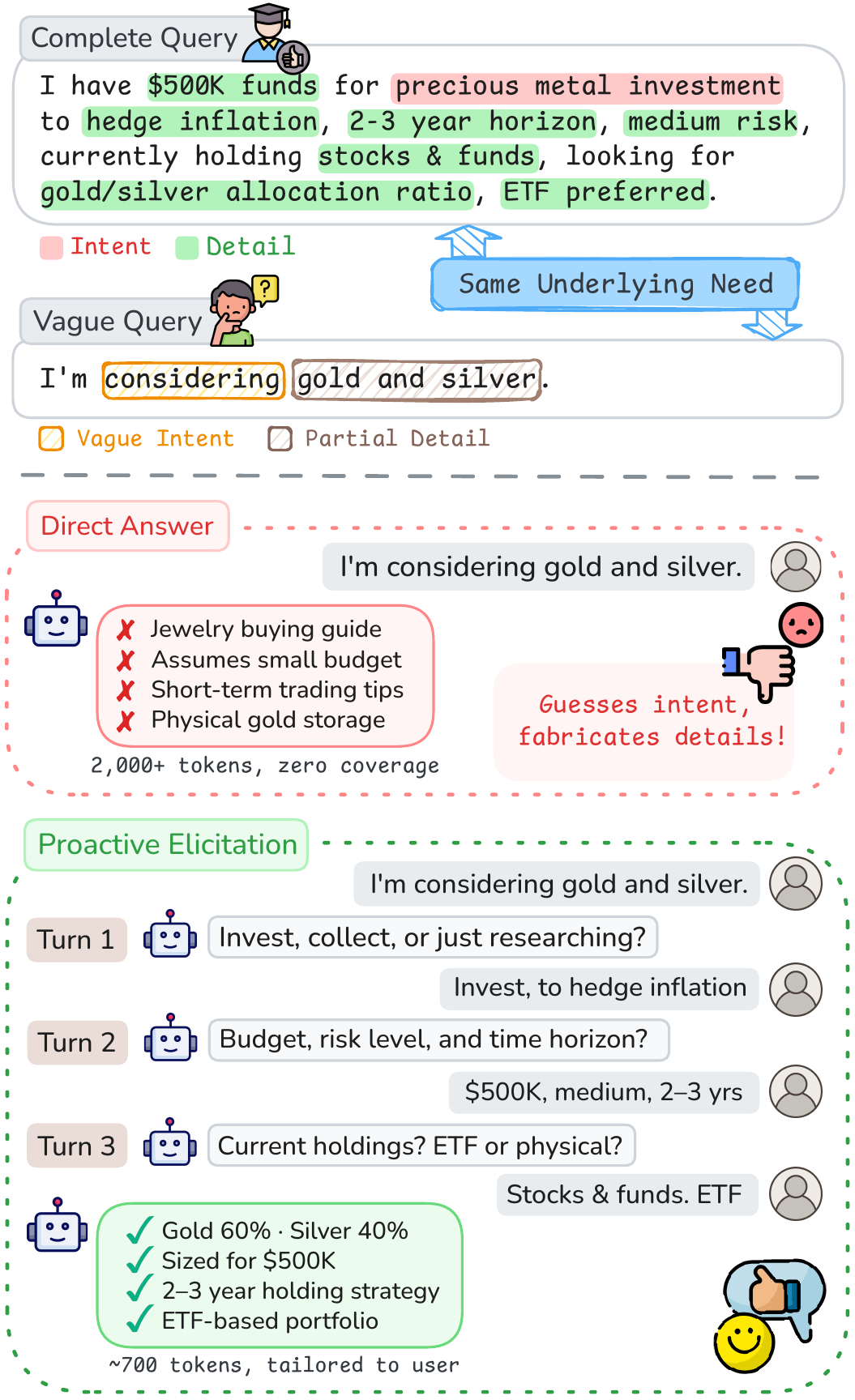} 
    \caption{
    \textbf{Top:} The same underlying need expressed as a complete query and a fully-vague query.
    \textbf{Bottom:} A direct-answer LLM fabricates all missing information, while proactive elicitation recovers it in 3 targeted turns.
    }
    \label{fig:motivation}
\end{figure}

\section{Introduction}
\label{sec:intro}
Today's large language models (LLMs) default to \emph{answering}, not \emph{asking}. When user queries are incomplete or underspecified, models silently infer unstated intent and constraints instead of asking targeted questions to resolve the gap (Figure~\ref{fig:motivation}). Yet real-world user queries are rarely self-contained: information gaps are pervasive in natural human communication \citep{taylor1967question}, and users routinely issue underspecified queries, overestimating the model's ability to infer unstated needs: task accuracy drops by 39\% when instructions are fragmented \citep{laban2025llms}, and medical diagnostic accuracy plummets from 94.9\% to 34.5\% when symptoms are underspecified\citep{bean2026reliability}.

Our analysis of 5{,}000 real-world queries from a commercial AI assistant further quantifies this gap: \textit{over half of user queries lack critical information, yet models from 8B to over 120B ask clarification in fewer than 11\% of such cases}, with no clear improvement from scaling.
These findings motivate \emph{proactive elicitation}: identifying missing information and asking targeted questions \emph{before} committing to an answer.

This capability is challenging because the information gap arises from both sides of the conversation. On the \textbf{user side}, users often leave critical constraints implicit, assuming the model shares context it does not have, a cognitive bias known as the \emph{curse of knowledge} \citep{camerer1989curse}. 
On the \textbf{model side}, post-training pipelines dominated by single-turn supervision and rewards for direct answers implicitly penalize pausing to ask. We formalize this gap at two levels: \textbf{(1)~Intent ambiguity}, where the user's overarching goal is unclear---e.g., whether ``help me work on this text'' means translation or summarization; and \textbf{(2)~Detail absence}, where the goal is clear but critical parameters are missing---e.g., budget or use case in ``recommend a laptop''. These two levels require different elicitation strategies and evaluation criteria.

Recent work has explored LLM clarification behavior \citep{shaikh2025navigating, zhang2025modeling, wu2025collabllm, andukuri2024star}, but remains limited along three axes. \textbf{Scope:} prior work mostly targets \emph{disambiguation}, resolving queries with multiple valid interpretations \citep{min2020ambigqa, zhang2024clamber}, rather than the broader case where essential information was \emph{never expressed} at all. \textbf{Data:} benchmarks are confined to narrow domains \citep{gan2409clarq} or synthetic ambiguity \citep{luo2025clarifymt}, with no training-ready resource grounded in real conversations. \textbf{Training:} methods rely on online user simulators or trajectory-level feedback \citep{wu2025collabllm, huang2025teaching, andukuri2024star, zhang2025modeling}, limiting fine-grained, per-step credit assignment.
 
To address these gaps, we propose a unified framework for evaluating and training proactive elicitation, grounded in real user--model conversations.
On the evaluation side, we introduce \textbf{VagueBench}, a benchmark built from real open-domain queries through \emph{controlled degradation}: intent-bearing linguistic cues are systematically weakened and task details removed, yielding vague variants whose hidden information set is precisely known. A simulated user holds the complete query and reveals a hidden information unit only when the model asks a semantically matching question, providing a controlled, reproducible testbed for multi-turn elicitation. We measure three complementary aspects: \emph{decision appropriateness}, whether the model asks rather than prematurely answers; \emph{elicitation efficiency}, the number of hidden information units recovered per turn; and \emph{answer coverage}, the F1 between the final response and the gold information set.
On the training side, we propose \textbf{DialogGRPO}, an offline reinforcement learning framework that models multi-turn elicitation as a dialogue-state directed acyclic graph (DAG). Each node groups alternative elicitation actions for local GRPO \citep{shao2024deepseekmath} comparison, and each edge receives a dense reward combining immediate information gain with long-term lookahead value, enabling the model to learn \emph{when} and \emph{what} to ask without the prohibitive cost of runtime user simulation.

Our contributions are threefold:

\begin{itemize}[leftmargin=*, itemsep=0pt, topsep=0pt, parsep=1pt]

\item \textbf{Benchmark.} We introduce VagueBench, a benchmark built
from real open-domain queries with controlled vagueness and known
hidden information for evaluating proactive elicitation capabilities
of LLMs.

\item \textbf{Method.} We propose DialogGRPO, which models multi-turn elicitation as a dialogue-state DAG and applies group-relative advantage estimation at each state for dense offline credit assignment.

\item \textbf{Findings.} Evaluation on VagueBench across open-source and
closed-source LLMs (8B to frontier) reveals systematic elicitation
suppression at all scales. DialogGRPO outperforms prompt-engineering and test-time-scaling baselines on all metrics, and zero-shot transfers to MultiWOZ 2.4 and AmbigQA.

\end{itemize}


\section{Related Work}
\label{sec:related}

\subsection{Proactive Elicitation}

Clarification and mixed-initiative interaction are well studied in conversational search and task-oriented dialogue\citep{budzianowski2018multiwoz, ye2022multiwoz, gan2409clarq}, where systems ask follow-up questions to resolve facets, slots, or preferences \citep{aliannejadi2019asking, aliannejadi2020convai3}. For LLM assistants, the challenge is broader: open-ended requests often omit intent or constraints, yet instruction-tuned models tend to answer directly and fill gaps by assumption \citep{yang2025prompts, laban2025llms}. Recent LLM work explores clarification via prompting, grounding analysis, and collaborative agents \citep{shaikh2025navigating, wu2025collabllm}, but mainly targets multiple valid interpretations. We instead study open-domain elicitation of unspoken intent and task details.

\subsection{Vagueness Benchmarks}

Existing benchmarks capture different forms of underspecification. Qulac and ClariQ evaluate clarification in conversational search \citep{aliannejadi2019asking, aliannejadi2020convai3}; AmbigQA and related datasets target questions with multiple valid interpretations \citep{min2020ambigqa, zhang2024clamber}; recent LLM benchmarks further test clarification in task-oriented dialogue, tool use, or synthetic ambiguity settings \citep{zhang2024clamber, gan2409clarq, luo2025clarifymt}. However, they assume a predefined domain, a fixed slot schema, or
a small set of alternative interpretations, providing limited coverage
of real open-domain vagueness where intent and task details may have
never been stated. VagueBench fills this gap with real assistant queries, unrestricted intents, and a multi-turn evaluation protocol.

\subsection{Training Elicitation Policies}

Prior methods improve clarification behavior through prompting, supervised fine-tuning, preference learning, simulated users, or downstream task rewards \citep{chen2025learning, andukuri2024star, zhang2025modeling, huang2025teaching}. Table~\ref{tab:related} compares these methods by whether they target open-domain interactions, use real user queries, avoid online user simulation, and provide turn-level rewards\citep{wu2025collabllm, mtpo}. While effective, existing approaches often depend on synthetic interactions, domain-specific settings, or coarse dialogue-level feedback. DialogGRPO instead precomputes a dialogue-state DAG and rewards each asking turn using both acquired information and downstream answer value.

\newcommand{\cmark}{\textcolor{green!55!black}{\ding{51}}}
\newcommand{\xmark}{\textcolor{red!75!black}{\ding{55}}}
\newcommand{\pmark}{\raisebox{0.15ex}{\scriptsize$\triangle$}}

\begin{table}[t]
\centering
\small
\resizebox{\columnwidth}{!}{%
\begin{tabular}{l cccc}
\toprule
\textbf{Method}
& \makecell[c]{\scriptsize\textbf{No Runtime}\\\scriptsize\textbf{Sim.}}
& \makecell[c]{\scriptsize\textbf{Open}\\\scriptsize\textbf{Domain}}
& \makecell[c]{\scriptsize\textbf{Turn}\\\scriptsize\textbf{Reward}}
& \makecell[c]{\scriptsize\textbf{Real}\\\scriptsize\textbf{Queries}} \\
\midrule
ClarifyGPT      & \cmark & \xmark & \xmark & \xmark \\
ACT             & \xmark & \xmark & \xmark & \xmark \\
STaR-GATE       & \cmark & \cmark & \xmark & \xmark \\
CollabLLM       & \xmark & \cmark & \cmark & \xmark \\
\rowcolor{gray!12}
\textbf{Ours}       & \cmark & \cmark & \cmark & \cmark \\
\bottomrule
\end{tabular}%
}
\caption{Comparison of proactive elicitation methods. No Runtime Sim.: the parameter-update phase does not invoke an LLM-based user simulator at runtime. Turn Reward: per-turn rather than trajectory-level credit assignment.}
\label{tab:related}
\end{table}


\begin{figure*}[t]
    \centering
    \includegraphics[width=\textwidth]{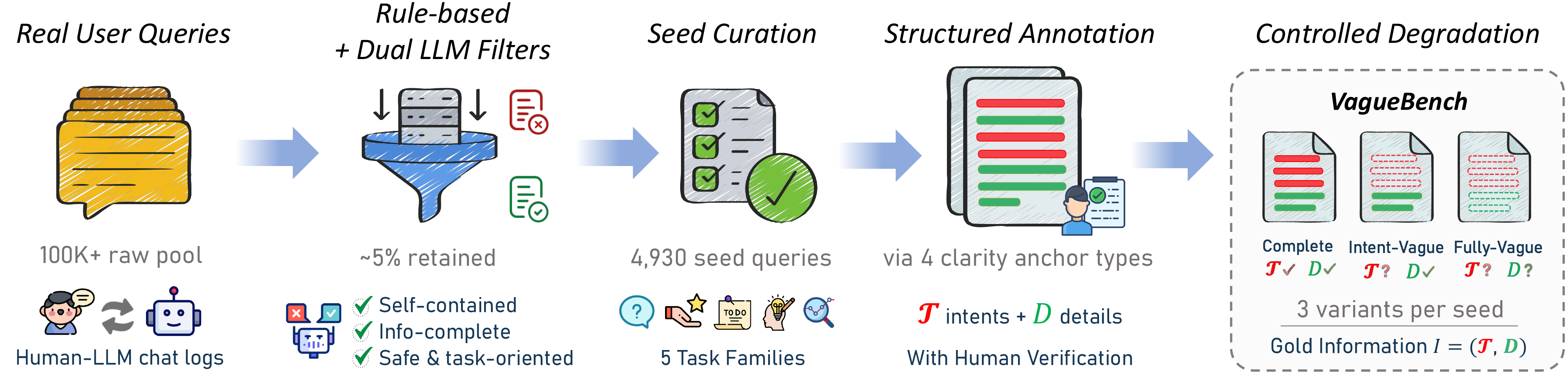} 
    \caption{VagueBench construction pipeline.
Real user queries are filtered for self-containedness and information completeness, annotated with intent and detail units via four clarity anchor types, and then systematically degraded into intent-vague and fully-vague variants with known gold information for evaluation.}
    \label{fig:vaguebench_pipeline}
\end{figure*}

\section{VagueBench}
\label{sec:benchmark}

A user submits a query $q_0$ that may express only part of a complete need $q^{\star}$.
We decompose $q^{\star}$ into an \emph{intent set} $\mathcal{T}=\{I_j\}_{j=1}^{m}$ of task-level goals and a \emph{detail set} $\mathcal{D}=\{d_i\}_{i=1}^{n}$ of key--value constraints for executing those goals; the gap between $q_0$ and $q^{\star}$ may involve
\emph{intent ambiguity}, \emph{detail absence}, or both,
as defined in \S\ref{sec:intro}.
Each benchmark instance begins with an information-complete seed whose $(\mathcal{T},\mathcal{D})$ is fully annotated, from which we derive an \emph{intent-vague} and a \emph{fully-vague} variant by progressively removing information units, so every gap is precisely quantifiable (Figure~\ref{fig:vaguebench_pipeline}).

\subsection{Data Collection}
\label{sec:data_collection}

We source over 100{,}000 open-domain queries from a commercial AI assistant.\footnote{From users who opted into data collection policy, with PII removed. See the Ethics Considerations.}
A rule-based pass enforces length bounds and removes unsafe content; two LLM passes then discard greetings, test strings, context-dependent utterances, and queries too detail-sparse for meaningful degradation.
The surviving 4{,}930 seeds ($\approx$5\,\%) span five task families: \emph{Information Query}, \emph{Recommendation \& Decision}, \emph{Task Execution}, \emph{Creation \& Generation}, and \emph{Consultation \& Analysis}, with the distribution shown in Figure~\ref{fig:query_distribution}.
Each seed carries on average $\bar{m}=$ 1.36 intents and $\bar{n}=$ 4.63 details.
Filtering rules and full statistics are in Appendix~\ref{sec:appendix_construction}.

\subsection{Data Construction}
\label{sec:data_construction}

The annotation scheme and variant construction share a single backbone: \emph{clarity anchors}, recurring linguistic cues that make a query's information explicit.

\paragraph{Clarity anchors.}
Drawing on argument-structure theory \citep{levin2005argument}
and speech-act theory \citep{searle1969speech}, we identify four
anchor types in two layers.
Consider the query
``\emph{\uline{Help me} \uline{summarize} this article into
\dashuline{five bullet points} \dashuline{focusing on the economic
implications}}'' as a running example, where \uline{solid} and
\dashuline{dashed} underlines mark intent-layer and detail-layer
anchors, respectively.
The \textbf{intent layer} signals what the user wants done:
\emph{predicate anchors} are action verbs that pin the task type
(\uline{\emph{summarize}}); \emph{speech-act anchors} are directive
frames (\uline{\emph{Help me\,\ldots}}).
The \textbf{detail layer} specifies how the task should be executed:
\emph{output-type anchors} mark the expected answer form
(\dashuline{\emph{five bullet points}}); \emph{scope anchors} bound
response dimensions
(\dashuline{\emph{focusing on economic implications}}).
This partition determines both what to annotate and what to remove.

\paragraph{Annotation.}
The anchor taxonomy makes annotation straightforward: intent-layer
anchors are grouped into task-level goals $\mathcal{T}$, while
detail-layer anchors and remaining surface constraints become
key--value pairs in $\mathcal{D}$, each linked to the intent it
serves.
In the running example, the predicate anchor
\uline{\emph{summarize}} and the speech-act anchor
\uline{\emph{Help me\,\ldots}} together identify a single
summarization intent.
On the detail side, the output-type anchor
\dashuline{\emph{five bullet points}} yields a key--value pair
$\langle\texttt{format},\;\text{five bullet points}\rangle$, and
the scope anchor \dashuline{\emph{focusing on the economic
implications}} yields
$\langle\texttt{focus},\;\text{economic implications}\rangle$;
both are linked to that summarization intent.
Two frontier LLMs (GLM-5) annotate each seed independently; disagreements
are resolved by a third LLM (GPT-5.4) acting as adjudicator, and a human
validation study (Appendix~\ref{sec:appendix_exp2}) confirms strong
agreement between LLM and human annotations ($\kappa \geq 0.81$).

\paragraph{Controlled degradation.}
Because anchors are partitioned into two layers, variant
construction is a principled removal procedure.
An \emph{intent-vague} variant weakens or removes only intent-layer
anchors while preserving every detail unit in $\mathcal{D}$.
A \emph{fully-vague} variant further strips detail-layer anchors
and a subset of detail units, retaining a core-topic phrase of
typically 5--10 tokens.  For example, the complete query
``\emph{\uline{Help me} \uline{summarize} this article into
\dashuline{five bullet points} \dashuline{focusing on the economic
implications}}'' reduces to
``\emph{Work on this article, \dashuline{five bullet points},
\dashuline{focusing on the economic implications}}'' at the
intent-vague level, and to
``\emph{Help me work on this article}'' at the fully-vague level.
Both variants undergo an LLM fluency rewrite; hidden-unit labels
are verified by set difference against the gold annotation, and
non-conforming cases are regenerated
(Appendix~\ref{sec:appendix_exp3}).

\subsection{Evaluation}
\label{sec:evaluation}

\paragraph{Interaction.}
A conversation begins with query $q_0$.
At each turn the model either asks a question or commits to a final
answer; the dialogue ends when the model answers, all hidden units
are recovered, or a budget of $T_{\max}$ turns is reached.
A simulated user $\mathcal{S}$, holding $q^{\star}$ and the hidden
units, reveals a unit only when the model's question semantically
matches it, releases no more than the matched units per turn, and
responds with a non-informative deflection to irrelevant or repeated questions. Details of simulated user is provided in Appedix~\ref{sec:appendix_simulated_user}.

\paragraph{Metrics.}
\noindent\textit{Decision appropriateness} (Ask\%): the first LLM output is classified as \emph{ask} or \emph{answer}; the gold label is \emph{ask} for vague variants and \emph{answer} for complete queries.

\noindent\textit{Elicitation efficiency} (IAR): the information acquisition rate measures hidden units recovered per elicitation turn, computed only when the model chose to ask:
\begin{equation}
\mathrm{IAR} = \frac{|\mathcal{T}_{\mathrm{acq}}|+|\mathcal{D}_{\mathrm{acq}}|}{\max(T,\,1)} .
\label{eq:iar}
\end{equation}

\noindent\textit{Answer coverage} (Cov.\ F1): the final answer is compared against the gold information set via F1 score. It is formulated as:
\begin{equation}
\mathrm{Cov.\ F1} = \frac{2 P_{\mathrm{cov}} R_{\mathrm{cov}}}{P_{\mathrm{cov}} + R_{\mathrm{cov}}}
\label{eq:cov_f1}
\end{equation}
where precision $P_{\mathrm{cov}}$ penalizes unsupported assumptions, and recall $R_{\mathrm{cov}}$ penalizes omissions.


\begin{figure*}[t]
    \centering
    \includegraphics[width=\textwidth]{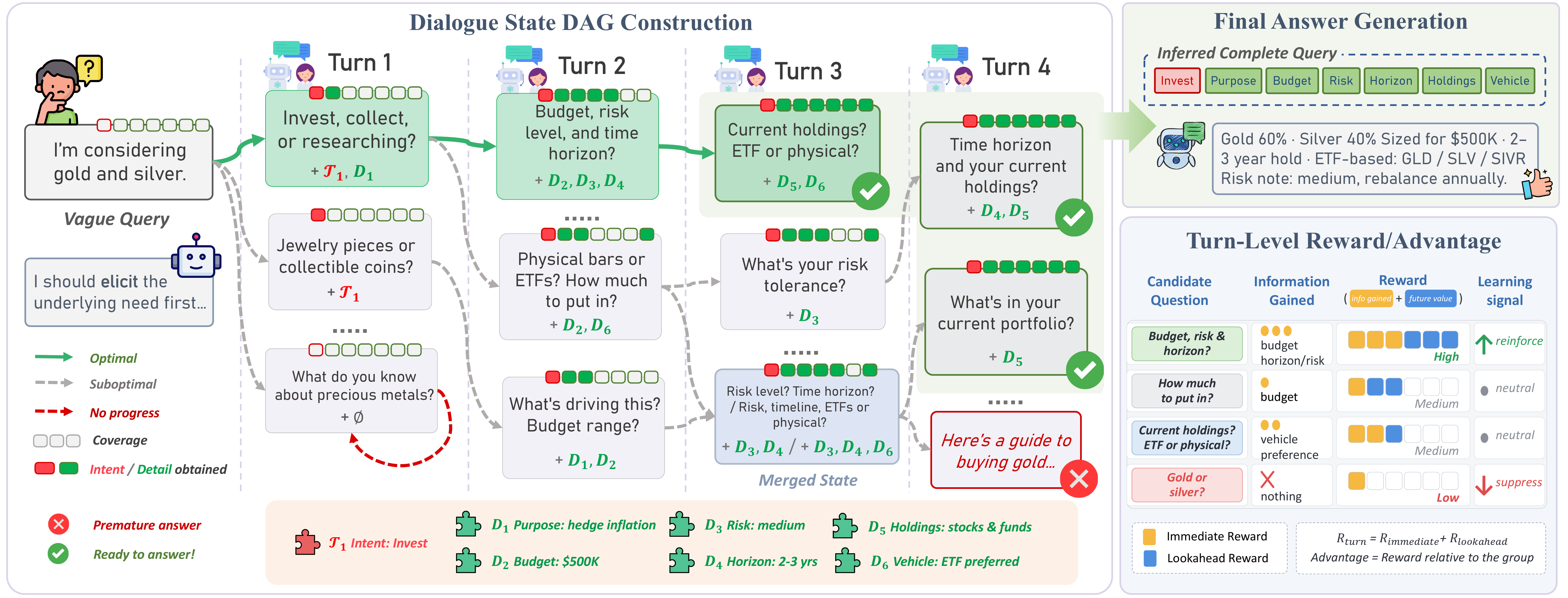} 
    \caption{
    DialogGRPO pipeline.
    \textbf{Left:} A dialogue-state DAG is built by expanding candidate
    questions at each turn and merging nodes with identical acquired
    sets.
    \textbf{Right:} Recovered units ground the final answer (top);
    turn-level rewards and group-relative advantages guide training
    (bottom).
    }
    \label{fig:dag_grpo}
\end{figure*}

\section{DialogGRPO}
\label{sec:method}

Training proactive elicitation faces two coupled challenges.
First, \emph{sparse credit assignment}: multi-turn rewards arrive only at the trajectory's end, leaving each intermediate question without direct feedback on whether it was worth asking.
Second, \emph{simulation cost}: on-policy rollouts require invoking an LLM-based user simulator at every gradient step, which is prohibitively expensive when each dialogue spans multiple turns.

DialogGRPO addresses both through a unified offline pipeline organized in three stages (Figure~\ref{fig:dag_grpo}).
We first construct a \emph{dialogue state DAG} (\S\ref{sec:dialogue_state_dag}) that maps out candidate elicitation questions and their consequences using an external LLM, so that the outgoing edges of any node form a natural GRPO comparison group.
We then assign each edge a \emph{turn-level reward} (\S\ref{sec:turn_reward}) that combines immediate information gain with lookahead value estimated from the pre-built graph, converting the single end-of-trajectory signal into dense per-action feedback.
Finally, we train the policy via \emph{advantage-weighted learning}
(\S\ref{sec:offline_grpo}) over (prompt, candidate, advantage) tuples
extracted from the DAG, requiring no runtime simulation.

\subsection{Dialogue State DAG}
\label{sec:dialogue_state_dag}

\paragraph{Nodes and termination.}
For a training query with information units $U^{\star}=\mathcal{T}\cup\mathcal{D}$, each node $v$ represents the acquired set $K_v\subseteq U^{\star}$, initialized at the root with units already present in $q_0$. A node is terminal when all units are acquired, an LLM answerability classifier deems $K_v$ sufficient for a satisfactory response, or the depth reaches $D_{\max}$. The classifier is a prompted LLM conditioned on $q_0$, $K_v$, and the dialogue history (Appendix~\ref{sec:appendix_prompts}).

\paragraph{Oracle-guided expansion.}
At each non-terminal node $v$, an external strong LLM (GLM-5.1), distinct from the policy model $\pi_\theta$, proposes $N$ candidate questions conditioned on $q_0$, the dialogue history reaching $v$, and the remaining gold units $U^{\star}\setminus K_v$.
The policy model never observes these gold units; the external LLM serves solely as an \emph{exploration engine} that populates diverse candidates. The policy learns to distinguish high-value questions through the group-relative advantage signal (\S\ref{sec:offline_grpo}), rather than through behavioral cloning or distillation of the external model.

For each candidate question $c$, the simulated user $\mathcal{S}$ produces a reply $u_c$, and an extraction function $\mathrm{Ext}(\cdot)$ maps the reply to a set of information units.
The \emph{information gain} of edge $e=(v,c)$ is:
\begin{equation}
\Delta_e =
\bigl(\mathrm{Ext}(u_c)\cap U^{\star}\bigr)\setminus K_v .
\label{eq:edge_delta}
\end{equation}
If $\Delta_e \neq \varnothing$, the edge leads to a child node $v'$ with $K_{v'}=K_v\cup\Delta_e$, and expansion continues from $v'$.
Candidates that acquire no new unit are recorded as \emph{null-progress} actions: they enter the comparison group at $v$ as negative examples but are not expanded further.

\paragraph{State merging.}
Different elicitation orders can reach the same acquired set.
We merge nodes sharing identical $K_v$, converting the expansion
tree into a compact DAG.
Merging avoids redundant expansion of duplicate states and allows
reward computations to be shared across paths that converge on the
same node.
When multiple dialogue histories converge at a merged node, we
retain the shortest-path history as the \emph{canonical prompt}
for that state; \S\ref{sec:offline_grpo} describes how this
interacts with training-tuple extraction.

\subsection{Turn-Level Reward}
\label{sec:turn_reward}

A root-to-terminal path $\tau=(e_1,\ldots,e_L)$ has a natural trajectory-level reward:
\begin{equation}
R_{\mathrm{traj}}(\tau)=F_1(v_L)-\lambda\,L ,
\label{eq:traj_reward}
\end{equation}
where $F_1(v_L)$ is the answer-coverage F1 at the terminal node and $\lambda$ penalizes interaction length.
However, this signal is available only at trajectory endpoints and cannot distinguish a good intermediate question from a bad one.
We therefore design two complementary per-edge signals.

\paragraph{Immediate information gain.}
Each edge $e$ receives:
\begin{equation}
R_{\mathrm{imm}}(e) = \alpha\,\min\!\bigl(|\Delta_e|,\;\delta_{\max}\bigr) ,
\label{eq:imm_reward}
\end{equation}
rewarding questions that elicit more units per turn.
The cap $\delta_{\max}$ prevents a single high-gain question from dominating the advantage distribution, keeping the reward discriminative across candidates of varying quality.
Null-progress edges receive $R_{\mathrm{imm}}=0$.

\paragraph{Lookahead value.}
The immediate reward is a \emph{single-step} signal that ignores
downstream consequences.
We complement it with a lookahead value estimated via Monte Carlo
rollouts in the pre-built DAG: for each edge $e$, we sample $K$
paths from its child state $v_e$ and average their trajectory
returns:
\begin{equation}
Q_{\mathrm{la}}(e)
= \tfrac{1}{K}\textstyle\sum_{k=1}^{K}
R_{\mathrm{traj}}(\tau_k\mid v_e) ,
\label{eq:lookahead}
\end{equation}
where $v_e$ is the child state reached by $e$.
For terminal $v_e$ we set $Q_{\mathrm{la}}(e)=F_1(v_e)$ directly.

\paragraph{Combined reward.}
The turn-level reward combines both signals:
\begin{equation}
R_{\mathrm{turn}}(e) = Q_{\mathrm{la}}(e) + R_{\mathrm{imm}}(e) .
\label{eq:turn_reward}
\end{equation}
For null-progress edges, we set $R_{\mathrm{turn}} = \bar{Q}_{\mathrm{la}}(v) - \lambda$, where $\bar{Q}_{\mathrm{la}}(v)$ is the mean lookahead at the parent node, reflecting that a null-progress question wastes one turn without narrowing the information gap between user and assistant.

\subsection{Advantage-Weighted Training}
\label{sec:offline_grpo}

\paragraph{Comparison groups and advantage estimation.}
For each non-terminal node $v$, the candidate set $\mathcal{G}_v=\{e:\mathrm{src}(e)=v\}$ forms a GRPO comparison group. A group contain different productive questions. Following the group-relative principle of GRPO~\citep{shao2024deepseekmath}, we compute a normalized advantage:
\begin{equation}
\hat{A}_e =
\frac{R_{\mathrm{turn}}(e)-\mu_v}{\sigma_v+\epsilon} ,
\label{eq:advantage}
\end{equation}
where $\mu_v$ and $\sigma_v$ are the within-group mean and standard deviation of $R_{\mathrm{turn}}$.

\paragraph{Training objective.}
Each training instance is a tuple $(p_v, y_e, \hat{A}_e)$, where $p_v$ is the canonical prompt at node $v$ (\S\ref{sec:dialogue_state_dag}) and $y_e$ is the candidate response associated with edge $e$.
We optimize:
\begin{equation}
{\small
\mathcal{J}(\theta) = \mathbb{E}_{v}\bigg[
\frac{1}{|\mathcal{G}_v|}
\sum_{e\in\mathcal{G}_v}
\exp\!\Big(\tfrac{\hat{A}_e}{\beta}\Big)\,
\log\pi_{\theta}(y_e\mid p_v)
\bigg]}
\label{eq:objective}
\end{equation}
where $\beta$ controls the strength of the KL penalty toward the
reference policy $\pi_{\mathrm{ref}}$.
Because all candidates are pre-generated and fixed in our offline
setting, there are no importance-sampling ratios or clipping\citep{ppo}, and the
objective reduces to \emph{advantage-weighted
regression}~\citep{peng2019advantage}.
A full derivation from the KL-regularized policy optimization
objective to Eq.~\ref{eq:objective} is provided in
Appendix~\ref{sec:appendix_derivation}.


\section{Experiments}
\label{sec:experiments}

\subsection{Setup}
\label{sec:setup}
The VagueBench test set comprises 250 complete queries and their
\emph{intent-vague} and \emph{fully-vague} variants (750 instances); the
remaining 4{,}680 seeds and their DAG expansions form the
training set.
For cross-dataset generalization we report on
MultiWOZ~2.4~\citep{ye2022multiwoz} and
AmbigQA~\citep{min2020ambigqa}, neither seen during training.
We evaluate seven LLMs spanning open-source
(Qwen3-8B, Llama-3.1-8B-Instruct, Qwen3.5-27B, GPT-oss-120B) and closed-source
(GPT-5.4, Claude Sonnet~4.6, Gemini~3.1~Pro) families.
Using Qwen3.5-27B as the backbone, we compare four methods:
\textbf{Prompt}, an elicitation-encouraging system prompt;
\textbf{UoT}~\citep{uot2024}, a test-time-scaling approach;
\textbf{SFT}, trained on highest-reward DAG trajectories; and our
\textbf{DialogGRPO}.
All dialogues are capped at $T_{\max}{=}10$ turns.
Implementation details are in Appendix~\ref{sec:appendix_implementation}.

\begin{table*}[t!]
\centering
\small
\setlength{\tabcolsep}{10pt}
\renewcommand{\arraystretch}{1.1}
\begin{tabular}{l ccc ccc}
\toprule
\multirow{2}{*}{\textbf{Model}}
& \multicolumn{3}{c}{\textbf{Intent-Vague}}
& \multicolumn{3}{c}{\textbf{Fully-Vague}} \\
\cmidrule(lr){2-4} \cmidrule(lr){5-7}
                     & Ask\% ($\uparrow$) & Cov. F1 ($\uparrow$) & IAR ($\uparrow$)
                    & Ask\% ($\uparrow$) & Cov. F1 ($\uparrow$) & IAR ($\uparrow$) \\
\midrule
\multicolumn{7}{l}{\textit{Closed-source Models}} \\
\midrule
GPT-5.4             & 1.6                 & 0.58                  & 0.06
                    & 10.8                & 0.53                  & 0.65   \\
Claude Sonnet 4.6   & 8.8      & \underline{0.60}                  & 0.08
                    & 8.4                 & 0.52                  & 0.56   \\
Gemini 3.1 Pro      & 2.4                 & 0.57                  & 0.00
                    & 3.6                 & 0.54       & \underline{0.80}   \\
\midrule
\multicolumn{7}{l}{\textit{Open-source Models}} \\
\midrule
Qwen3-8B            & 1.2                 & 0.47                  & 0.00
                    & 1.6                 & 0.36                  & 0.21 \\
Llama-3.1-8B-Instruct & 0.8               & 0.50                  & 0.00
                    & 1.2                 & 0.35                  & 0.23 \\
Qwen3.5-27B         & 1.6                 & 0.49                  & 0.08
                    & 2.0                 & 0.38                  & 0.37 \\
GPT-oss-120B        & 2.0                 & 0.46                  & 0.02
                    & 3.6                 & 0.36                  & 0.35 \\
\midrule
\multicolumn{7}{l}{\textit{Methods (built on Qwen3.5-27B)}} \\
\midrule
\quad + Prompt      & 38.4                & 0.52                  & 0.22
                    & 47.2                & 0.50                  & 0.38 \\
\quad + UoT         & 58.8                & 0.50                  & 0.45
                    & 62.4                & 0.49                  & 0.41 \\
\rowcolor{gray!10}
\quad + SFT (Ours)
        & \underline{82.0}      & \underline{0.60}                & \underline{0.64}
        & \underline{85.6}      & \underline{0.63}                & \underline{0.80} \\
\rowcolor{gray!10}
\quad \textbf{+ DialogGRPO  (Ours)}
        & \textbf{88.4}            & \textbf{0.64}        & \textbf{0.76}
        & \textbf{91.2}            & \textbf{0.67}        & \textbf{0.87} \\
\bottomrule
\end{tabular}
\caption{
Main results on VagueBench.
\emph{Ask\%}: proportion of samples on which the model chose to ask.
\emph{Cov.\ F1}: coverage F1 of the final answer against the gold information set.
\emph{IAR}: information acquisition rate (Eq.~\ref{eq:iar}), computed only on samples where the model asked.
Best in \textbf{bold}; second best \underline{underlined}.
}
\label{tab:main}
\end{table*}

\subsection{Main Results}
\label{sec:main_results}



\noindent\textbf{Elicitation suppression persists across scales.}
As shown in Table~\ref{tab:main}, across all evaluated models, Ask\% on \emph{intent-vague} queries stays at or below $9\%$ and on \emph{fully-vague} queries never exceeds $11\%$.
This suppression holds regardless of model family or size: the four open-source models achieve Ask\% of only $0.8$--$2.0$ on intent-vague, while the three closed-source frontier models range from $1.6$ to $8.8$, with larger models showing no consistent advantage.
The shortfall suggests a post-training tendency to answer directly, rather than a capability limit that further scaling would fix.

\noindent\textbf{Prompting reveals latent elicitation ability but quickly plateaus.}
A single elicitation-encouraging prompt lifts Qwen3.5-27B Ask\% from $1.6\%$ to $38.4\%$ on \emph{intent-vague} and from $2.0\%$ to $47.2\%$ on \emph{fully-vague}, confirming that the ability resides in the pretrained weights, suppressed rather than absent.
However, the prompted model asks more often without asking more efficiently: IAR stays at $0.22$/$0.38$, far below the levels that training-based methods reach.
Prompting can shift the ask-versus-answer decision boundary but provides no signal for improving per-turn elicitation efficiency, motivating an approach that optimizes both \emph{when} and \emph{what} to ask.

\noindent\textbf{DialogGRPO asks better, not just more often.}
Compared with SFT trained on the highest-reward trajectories, DialogGRPO further improves Ask\% by up to $6.4$ points and IAR by $0.07$--$0.12$, confirming that group-relative advantage estimation provides a finer learning signal than trajectory-level imitation.
The Cov.\ F1 improvement over Qwen3.5-27B ($0.64$/$0.67$ vs.\ $0.49$/$0.38$) directly reflects the value of elicitation: the model answers on \emph{recovered} information rather than \emph{assumed} defaults.
UoT presents a different bottleneck: per-turn recovery is reasonable when it does ask (IAR $0.45$/$0.41$), but its uncertainty heuristic triggers elicitation on only $\sim$60\% of vague queries, leaving substantial information unrecovered.

\begin{table}[t]
\centering
\small
\setlength{\tabcolsep}{6pt}
\renewcommand{\arraystretch}{1.1}
\resizebox{\columnwidth}{!}{%
\begin{tabular}{l cc}
\toprule
\textbf{Model} & Ask\% ($\downarrow$) & Cov. F1 ($\uparrow$) \\
\midrule
Qwen3.5-27B      & \textbf{1.6}   & 0.58 \\
\quad + Prompt    & 42.8   & 0.59 \\
\quad + UoT          & 56.4   & 0.48    \\
\rowcolor{gray!10}
\quad + SFT (Ours)       & 6.4   & \underline{0.71} \\
\rowcolor{gray!10}
\quad \textbf{+ DialogGRPO  (Ours)} & \underline{5.2}   & \textbf{0.74}    \\
\bottomrule
\end{tabular}%
}
\caption{
Behavior on information-complete seed queries. Ask\% is lower-the-better.
Best in \textbf{bold}; second best \underline{underlined}.
}
\label{tab:complete}
\end{table}

\noindent\textbf{High elicitation on vague queries does not cause over-asking on complete ones.}
On information-complete queries (Table~\ref{tab:complete}), DialogGRPO asks on only $5.2\%$ of samples, close to Qwen3.5-27B's $1.6\%$, while lifting Cov.\ F1 from $0.58$ to $0.74$.
Prompt and UoT, by contrast, push Ask\% to $42.8\%$ and $56.4\%$, unnecessarily interrupting roughly half of already well-specified queries.
This calibration arises naturally from the group-relative advantage: when all candidate questions at a state yield uniformly low information gain, the advantage signal \emph{suppresses} asking without requiring an explicit threshold.

\noindent\textbf{General capability is preserved.}
Beyond elicitation metrics, we verify that DialogGRPO introduces no catastrophic forgetting relative to the untrained Qwen3.5-27B: on six benchmarks spanning coding, math, commonsense reasoning, and general knowledge, it matches or slightly exceeds the original model (Appendix~\ref{sec:appendix_general_capability}), indicating that offline advantage-weighted RL stays close enough to the reference policy to avoid capability regression.

\begin{table}[t]
\centering
\small
\setlength{\tabcolsep}{4pt}
\renewcommand{\arraystretch}{1.1}
\begin{tabular}{l cc cc}
\toprule
\multirow{2}{*}{\textbf{Variant}}
& \multicolumn{2}{c}{\textbf{Intent-Vague}}
& \multicolumn{2}{c}{\textbf{Fully-Vague}} \\
\cmidrule(lr){2-3} \cmidrule(lr){4-5}
                                & Cov. F1         & IAR          & Cov. F1        & IAR \\
\midrule
\rowcolor{gray!10}
\textbf{Full} (Ours)            & \textbf{0.64} & \textbf{0.76} & \textbf{0.67} & \textbf{0.87} \\
\midrule
\quad w/o lookahead reward        & 0.58          & \underline{0.74}          & 0.60          & \underline{0.84}          \\
\quad w/o immediate reward     & \underline{0.60}          & 0.68          & \underline{0.66}          & 0.85          \\
\bottomrule
\end{tabular}
\caption{
Reward ablation on VagueBench. The lookahead term primarily affects Cov.\ F1; the immediate term primarily affects IAR.
Best in \textbf{bold}; second best \underline{underlined}.
}
\label{tab:ablation}
\end{table}

\subsection{Ablation Study}
\label{sec:ablation}

Table~\ref{tab:ablation} ablates the two reward signals defined in \S\ref{sec:turn_reward}.
Without the lookahead term, Cov.\ F1 drops by $0.06$--$0.07$ while IAR remains nearly intact, suggesting that the policy becomes myopic and favors locally greedy questions over those that lead to higher final answer coverage.
Conversely, removing the immediate term leaves Cov.\ F1 largely unchanged but reduces IAR by up to $0.08$, as the model loses the per-turn incentive to pack multiple information units into a single question.
Neither ablated variant matches the full method, confirming that the two reward signals are complementary.

\begin{table}[t]
\centering
\small
\setlength{\tabcolsep}{5pt}
\renewcommand{\arraystretch}{1.15}
\resizebox{\columnwidth}{!}{%
\begin{tabular}{l c cc}
\toprule
\multirow{2}{*}{\textbf{Model}}
& \textbf{MultiWOZ 2.4}
& \multicolumn{2}{c}{\textbf{AmbigQA}} \\
\cmidrule(lr){2-2} \cmidrule(lr){3-4}
              & Success\% & Ask\% & Ask F1\% \\
\midrule
GPT-5.4                  & \textbf{39.9} & 5.1 & 9.6 \\
GLM-5                    & 23.8 & 0.9 & 1.9 \\
\midrule
Qwen3.5-27B              & 11.4 & 1.1 & 2.1 \\
\rowcolor{gray!10}
\quad \textbf{+ DialogGRPO (Ours)} & \underline{24.7} & \underline{16.3} & \underline{26.3} \\
\midrule
Gemma3-27B               & 15.1 & 1.1 & 2.2 \\
\rowcolor{gray!10}
\quad \textbf{+ DialogGRPO (Ours)} & 17.8 & \textbf{34.4} & \textbf{44.7} \\
\bottomrule
\end{tabular}%
}
\caption{
Zero-shot transfer. \emph{Success\%}: MultiWOZ task-completion rate. \emph{Ask\%}: fraction of queries on which the model asks in the AmbigQA \emph{multipleQAs} subset. \emph{Ask F1\%}: ask-versus-answer F1 on the full AmbigQA development set.
Best in \textbf{bold}; second best \underline{underlined}.
}
\label{tab:generalization}
\end{table}

\subsection{Generalization}
\label{sec:generalization}

\paragraph{Across datasets.}
To test whether the learned elicitation ability overfits to VagueBench, we evaluate zero-shot on two out-of-distribution benchmarks: MultiWOZ~2.4, a task-oriented slot-filling dialogue dataset, and AmbigQA, an ambiguous question-answering dataset. We train DialogGRPO on Qwen3.5-27B and Gemma3-27B and compare their performance against two frontier models (GPT-5.4 and GLM-5). For MultiWOZ, we follow the zero-shot end-to-end simulated interaction protocol of \citet{speakrl}. For AmbigQA, Ask\% is the fraction of queries on which the model asks in the ambiguous \emph{multipleQAs} subset. Ask F1\% is computed on the full development set with \emph{Ask} as the positive class, assigning gold \emph{Ask} to ambiguous queries and gold \emph{Answer} to unambiguous queries.

On AmbigQA, the benchmark most aligned with the elicitation capability we train, both models exhibit substantial gains (Table~\ref{tab:generalization}). Notably, Gemma3-27B is an instruction-tuned model while Qwen3.5-27B is a reasoning model, the elicitation behavior therefore transfers consistently across two distinct model types. On MultiWOZ, Success\% is a task-completion metric bounded by the base model's downstream execution rather than elicitation alone, yet both models still achieve clear improvements.
The consistent gains across slot-filling dialogue and ambiguous QA suggest that proactive elicitation is a transferable capability: one that is suppressed by default but recoverable through targeted training with dense credit assignment.

\begin{figure}[t]
\centering
\includegraphics[width=\columnwidth]{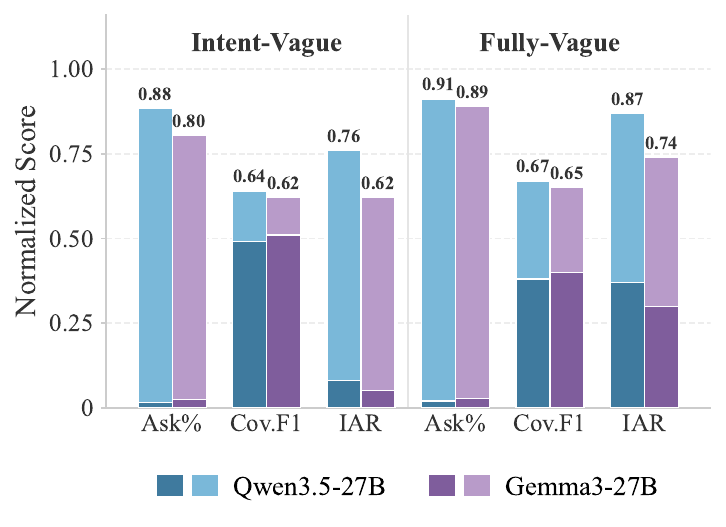}
\caption{Generalization across base models on VagueBench. Darker/lighter segments denote base model scores and DialogGRPO lifts, respectively. Both 27B models show comparable gains, confirming transferability across model families.}
\label{fig:cross_base}
\end{figure}

\paragraph{Across base models.}
To verify that the gains stem from the training recipe rather than from a particular base model, we re-ran the full DialogGRPO pipeline on Gemma3-27B with identical hyperparameters (Appendix~\ref{sec:appendix_gemma}).
Before training, both Qwen and Gemma 27B models exhibit the same suppression pattern (Ask\% below $3\%$ on both splits).
After training, both converge to comparable Ask\%, Cov.~F1, and IAR, with similarly sized lifts on every metric (Figure~\ref{fig:cross_base}), confirming that DialogGRPO generalizes across model families.

\section{Conclusion}

We introduce VagueBench, a benchmark exposing that elicitation suppression in LLMs persists across model families and scales, and DialogGRPO, an offline RL method that recovers this latent ability via dialogue-state DAG construction and dense turn-level credit assignment. DialogGRPO teaches models to ask better questions, not merely more, reaching 88--91\% Ask\% on vague queries while preserving calibration on complete ones, and generalizes zero-shot to MultiWOZ~2.4 and AmbigQA.

\section*{Limitations}

VagueBench is built from a single commercial assistant's traffic, which limits its coverage of conversational scenarios, languages, and modalities.
Our evaluation relies on a simulated user that cooperatively releases information on matching questions, abstracting away real human behaviors such as partial answers, sudden topic switches, noise, or refusal to elaborate, so the reported gains likely represent an upper bound on what can be achieved with cooperative interlocutors.
Open domain dialogue spaces are unbounded, so the DAG covers only a sparse set of trajectories. Local contrastive coverage and state merging reduce this gap, but interactions outside the training distribution remain a limitation.
Large preference spaces also make exhaustive questioning impractical. In shopping, a few targeted questions can narrow the options, but preferences may emerge or change during the interaction. VagueBench fixes the hidden preference set for reproducibility and does not capture this dynamic. Asking frequency is controlled by reward hyperparameters and should be calibrated to application needs.

\section*{Ethical Considerations}

We have conducted this work with a strong commitment to ethical and responsible research practices. The raw queries used to construct VagueBench originate from a commercial AI assistant that offers tiered data-sharing preferences, only interactions from users who explicitly opted into research use were included. All queries were stripped of personally identifiable information prior to our access, and a dedicated unsafe-content filter removed harmful or sensitive material during the rule-based filtering stage (\S\ref{sec:data_collection}). The released benchmark contains only post-filtering, annotation-enriched seeds with no metadata linking back to individual sessions. For the human validation study, two graduate student annotators were recruited and compensated at \$7.50 USD per hour; given the flexibility of the remote task and the economic context of the participants' residential region, this rate exceeds the local average hourly wage, ensuring fair remuneration. Participation was voluntary, and annotators provided informed consent for the use of their labels. Proactive elicitation is inherently dual-use: it reduces unfounded assumptions and prevents costly errors in high-stakes domains, but could also be steered into over-soliciting personal information unnecessary for the task. Our reward formulation mitigates this by crediting only information units causally required for the final answer; nevertheless, we recommend that production deployments constrain what the model may solicit, audit questions touching sensitive attributes, and give users a clear option to decline. We also note that VagueBench covers English-only queries from a single platform, limiting its linguistic and cultural generalizability. All data released are for research purposes and recommend that any production use be paired with appropriate guardrails.

\section*{Acknowledgments}

This work was supported by the National Natural Science Foundation of China under Grants 42394060 and 42394064, and the Ant Group Research Fund.

\bibliography{custom}

\clearpage
\appendix
\section{Data Construction and Statistics}
\label{sec:appendix_construction}

This appendix expands on \S\ref{sec:data_construction} with the
filtering pipeline that produces the 4{,}930-seed pool and four
statistical views of that pool: its task-family composition, the
per-seed counts of annotated intents and details, and the initial
information coverage of the two vague variants. Together they
characterise the bench used throughout the main text.

\begin{figure*}[t]
\centering
\includegraphics[width=\linewidth]{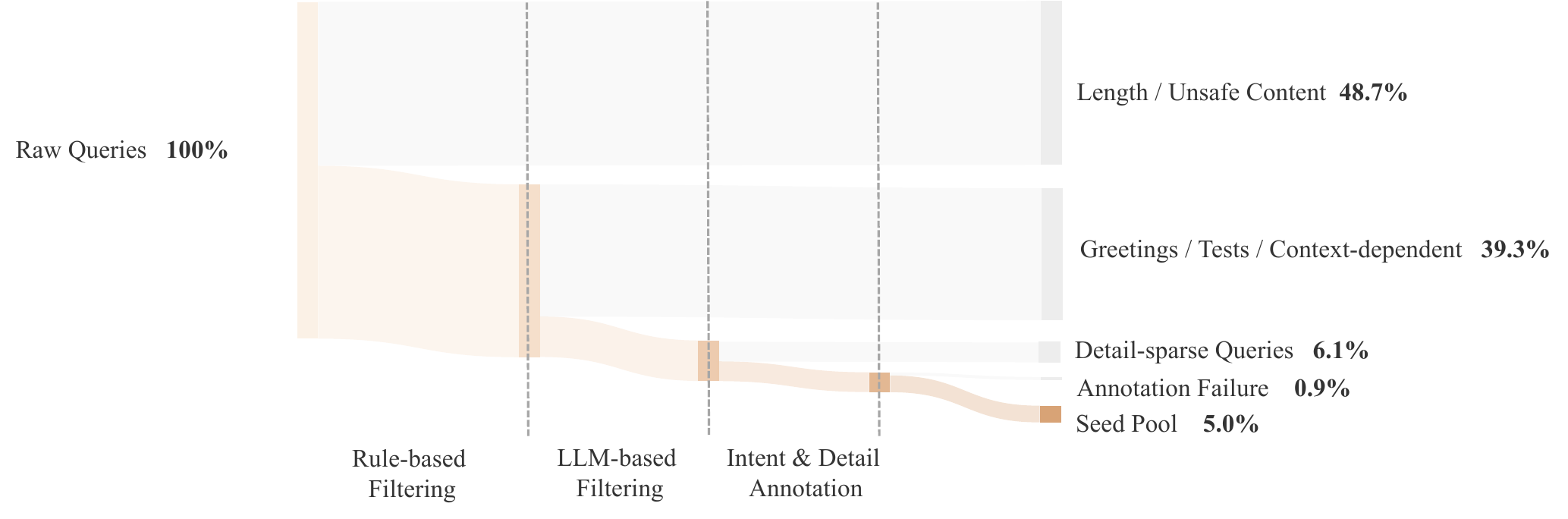}
\caption{
Flow of queries through the four-stage VagueBench filtering pipeline.
\emph{Rule-based Filtering} removes $48.7\%$ of the $\sim$100K raw
queries on length or unsafe-content grounds. \emph{LLM-based
Filtering} then discards greetings, test strings, and context-dependent
utterances ($39.3\%$), followed by detail-sparse queries that lack the
structure required for degradation ($6.1\%$). A final \emph{Intent \&
Detail Annotation} stage yields $4{,}930$ information-complete seeds,
$5.0\%$ of the input. The lighter peach tones on the left mark earlier
stages and the deepest tone on the right marks the final seed pool;
grey side-flows are the discarded branches at each stage.
}
\label{fig:cleaning_pipeline}
\end{figure*}

\paragraph{Filtering pipeline.}
Figure~\ref{fig:cleaning_pipeline} traces the flow of queries through
the four-stage filtering pipeline described in
\S\ref{sec:data_collection}. \emph{Rule-based Filtering} on length and
unsafe content removes $48.7\%$ of raw queries. Two subsequent stages
of \emph{LLM-based Filtering} strip greetings, test strings, and
context-dependent utterances ($39.3\%$), then queries that are too
detail-sparse to degrade meaningfully ($6.1\%$). A final
\emph{Intent \& Detail Annotation} stage rejects another $0.9\%$ for
annotation failure, leaving $5.0\%$ of the original $\sim$100K queries
as the information-complete seed pool used throughout the paper.

\begin{figure}[t]
\centering
\includegraphics[width=\linewidth]{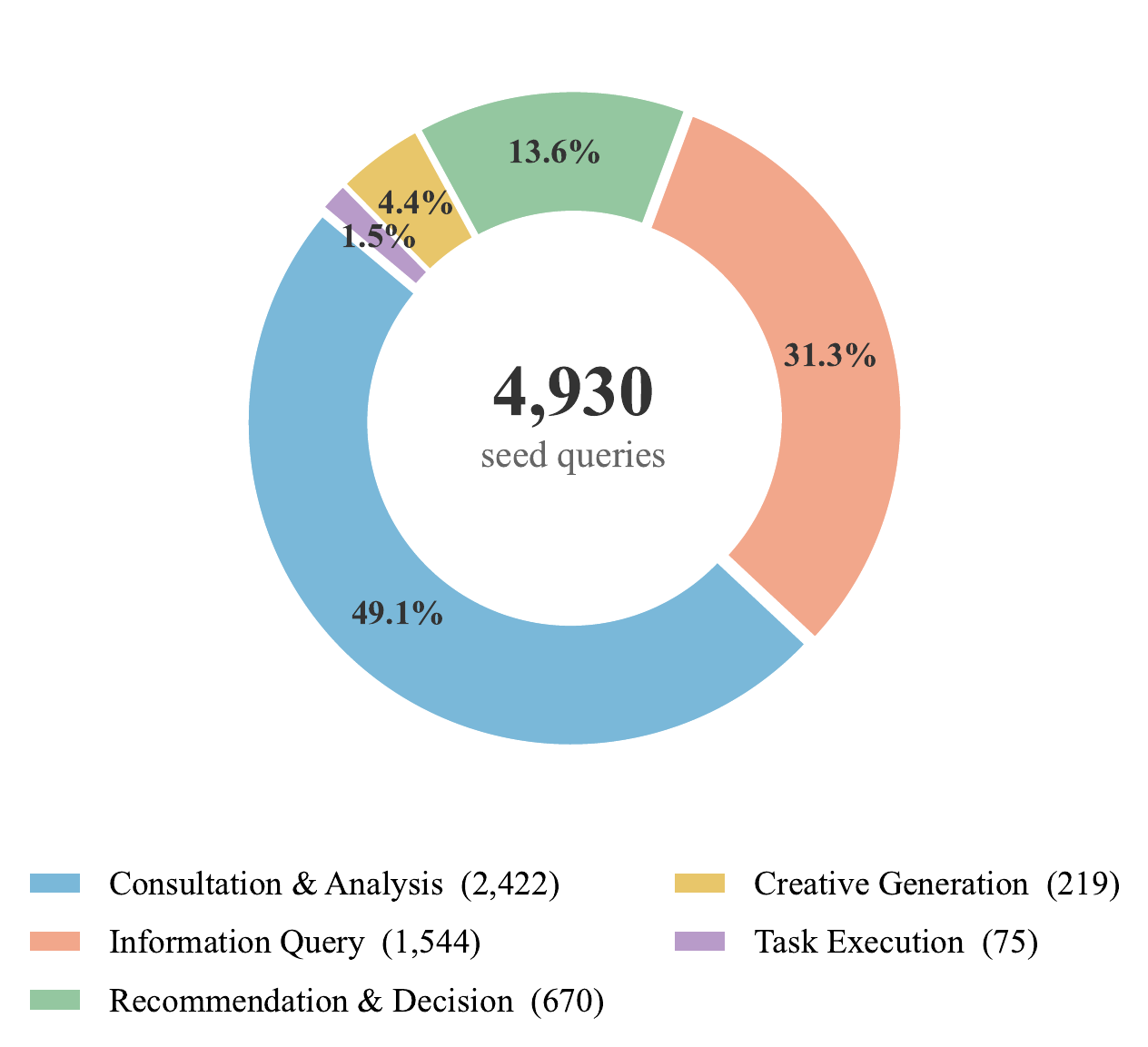}
\caption{
Task-family composition of the $4{,}930$ VagueBench seed queries.
\emph{Consultation \& Analysis} and \emph{Information Query} together
account for over $80\%$ of the pool, while \emph{Task Execution} forms
the rarest $1.5\%$. The bench inherits a long-tailed mix that mirrors
real-world assistant traffic.
}
\label{fig:query_distribution}
\end{figure}

\paragraph{Task-family distribution.}
Figure~\ref{fig:query_distribution} reports the composition of the
$4{,}930$-seed pool by task family. \emph{Consultation \& Analysis} and
\emph{Information Query} dominate at $49.1\%$ and $31.3\%$ respectively,
with \emph{Recommendation \& Decision} queries forming a further $13.6\%$.
\emph{Creation \& Generation} and \emph{Task Execution} form the long
tail at $4.4\%$ and $1.5\%$. The distribution inherits the empirical mix
observed in the source assistant traffic.

\begin{figure*}[t]
\centering
\includegraphics[width=\textwidth]{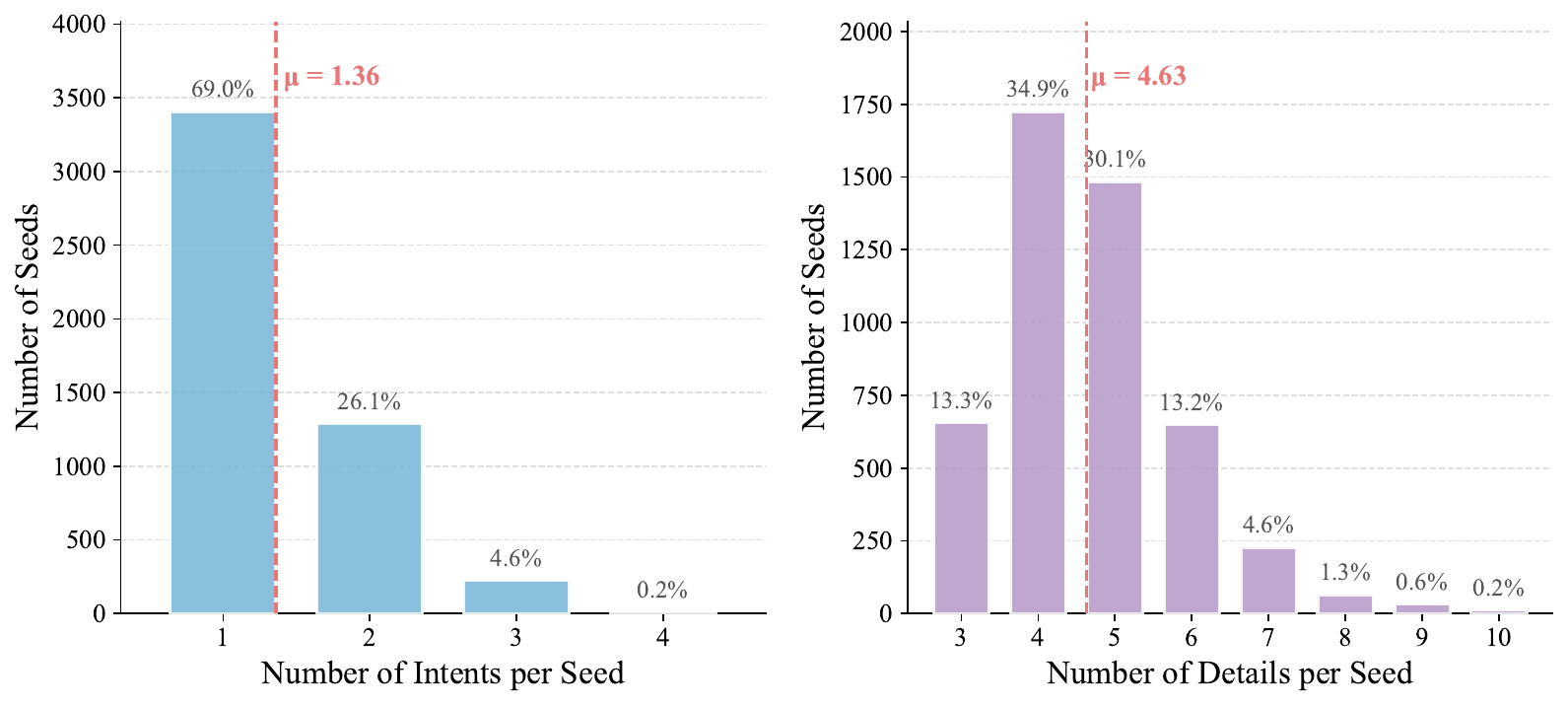}
\caption{
Per-seed counts of annotated intents (left) and details (right) in the
$4{,}930$-seed pool. Seeds carry on average $1.36$ intents and $4.63$
details; $69.0\%$ have exactly one intent, and detail counts cluster in
the 3 to 6 range.
}
\label{fig:intent_detail_distribution}
\end{figure*}

\paragraph{Information density of seeds.}
Figure~\ref{fig:intent_detail_distribution} shows the per-seed distribution
of annotated intents and details. VagueBench is anchored in single-intent
queries, with $69.0\%$ of seeds carrying exactly one intent and a mean of
$1.36$ intents per seed. Detail counts form a roughly bell-shaped
distribution peaking at four details per seed ($34.9\%$) with a mean of
$4.63$. This information density is what makes the bench non-trivial:
every seed carries multiple removable units, so the degraded variants
expose real hidden information rather than vacuous gaps.

\begin{figure}[t]
\centering
\includegraphics[width=\linewidth]{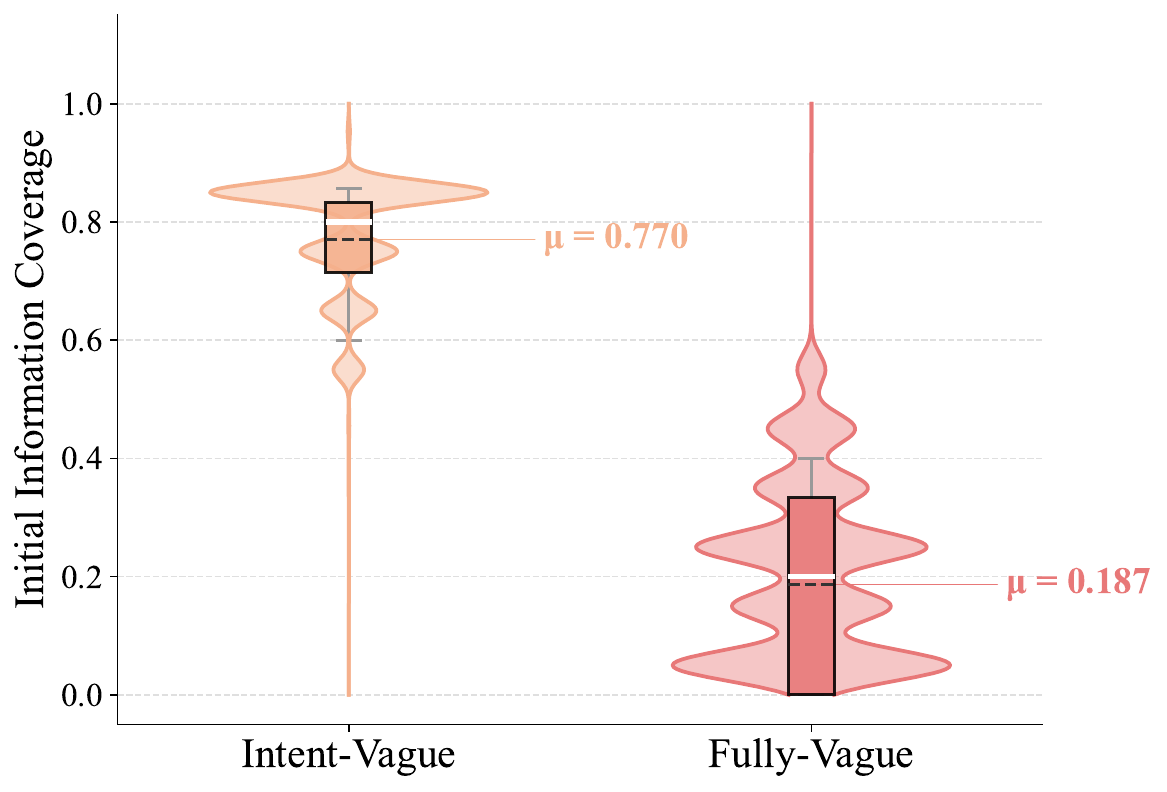}
\caption{
Initial information coverage of each query variant. Coverage is the
fraction of gold information units, intents and details combined, that
are already expressed in the initial query. Intent-Vague (peach)
preserves $77.0\%$ of units on average since only intent anchors are
weakened, while Fully-Vague (coral) drops to $18.7\%$ as both intent and
a sample of detail anchors are removed. KDE smoothed; box-and-whisker
overlays show median, interquartile range, and 5/95 percentiles.
}
\label{fig:coverage}
\end{figure}

\paragraph{Initial information coverage of the variants.}
Figure~\ref{fig:coverage} measures the initial information coverage of each
query variant, that is, the fraction of gold information units already
expressed in the initial query. Intent-Vague preserves $77.0\%$ of units on
average by construction, since it weakens only intent anchors; Fully-Vague
drops coverage to $18.7\%$, removing intent anchors together with a sample
of detail anchors. The two distributions barely overlap, confirming that
the controlled degradation produces two cleanly separated evaluation
regimes that correspond exactly to the two splits used in
\S\ref{sec:experiments}.

\section{Human Study}
\label{sec:appendix_human_study}

To validate three components of our pipeline, we conducted a small human
study with two annotators who were graduate students familiar with
dialogue research and blind to each other's labels. The three experiments
verify, in order, whether the vague queries in VagueBench genuinely
warrant clarification, whether the synthesized intent and detail
annotations align with human judgement, and whether the LLM extractor used
in our evaluation protocol recovers the same information units that a
human reader would. All agreement scores are reported as Cohen's $\kappa$
at the unit level.

\subsection{Experiment 1: Do Vague Queries Genuinely Warrant Clarification?}
\label{sec:appendix_exp1}

\noindent\textbf{Setup.}
We sampled $50$ queries uniformly at random from the Intent-Vague split
and $50$ from the Fully-Vague split, for a total of $100$ queries. To verify if the omitted details were genuinely necessary, the two annotators were shown the vague query alongside the complete gold query $q^{\star}$. Each annotator independently judged whether the information gap objectively warrants at least one clarifying question before attempting an answer, recording a binary Ask or Answer decision.

\noindent\textbf{Results.}
Table~\ref{tab:human_exp1} reports the two annotators' Ask rates on each
split. On Intent-Vague queries both annotators chose to ask on at least
$96\%$ of samples, and on Fully-Vague queries both chose to ask on
$100\%$. The results indicate that the controlled degradation in
VagueBench produces vague queries that human readers consistently
recognise as information-deficient, and that the gap between this human
Ask rate and the $1.6\%$ and $2.0\%$ Ask rate of un-prompted Qwen3.5-27B
in Table~\ref{tab:main} reflects a behavioural suppression in the model
rather than ambiguity in the data.

\begin{table}[htbp]
\centering
\small
\setlength{\tabcolsep}{8pt}
\renewcommand{\arraystretch}{1.15}
\begin{tabular}{l cc}
\toprule
\textbf{Split} & Human~1 Ask\% & Human~2 Ask\% \\
\midrule
Intent-Vague   & 98.0 & 96.0 \\
Fully-Vague    & 100.0 & 100.0 \\
\bottomrule
\end{tabular}
\caption{
Human Ask rates on vague queries. Both annotators were given the vague
query $q$ together with its complete gold query $q^{\star}$ and judged
whether the information gap warranted a clarifying question before
answering. High Ask rates confirm that the vague queries genuinely
warrant clarification.
}
\label{tab:human_exp1}
\end{table}

\subsection{Experiment 2: Are the LLM-Annotated Intent and Detail Units Accurate?}
\label{sec:appendix_exp2}

\noindent\textbf{Setup.}
We sampled $50$ complete queries $q^{\star}$ from the test pool. For each
query, the two human annotators and the LLM annotator (GLM-5) each independently
extracted the intent and detail units $\mathcal{T} \cup \mathcal{D}$ directly
from the query, with no access to one another's outputs. After this independent extraction, the human judges manually aligned the three annotation sets while viewing the complete query. They treated two units as identical when they denoted the same intent or the same detail key--value pair, regardless of surface wording. The union of all aligned units forms the per-query decision pool, on which each unit receives
a binary label from every annotator indicating whether that annotator
extracted it. Cohen's $\kappa$ is computed pairwise at the unit level, pooled
over all $50$ queries. Because the pool is the union of all three sources
rather than a pre-given candidate list, this protocol reflects both spurious
units and missing units, so the agreement scores capture annotation
completeness as well as precision.

\noindent\textbf{Results.}
Table~\ref{tab:human_exp2} reports the three pairwise Cohen's $\kappa$
values between the two annotators and the LLM annotation. Human~1 and
Human~2 reached $\kappa = 0.87$, indicating that the unit-level Present
or Absent criterion is well-defined. The agreement between each
annotator and the LLM annotation was $\kappa = 0.83$ and $\kappa = 0.81$,
about $0.05$ below the inter-annotator ceiling, supporting the use of
LLM-extracted annotations as gold labels in VagueBench.

\begin{table}[htbp]
\centering
\small
\setlength{\tabcolsep}{8pt}
\renewcommand{\arraystretch}{1.15}
\begin{tabular}{l c}
\toprule
\textbf{Pair} & Cohen's $\kappa$ \\
\midrule
Human~1 vs Human~2  & 0.87 \\
Human~1 vs LLM      & 0.83 \\
Human~2 vs LLM      & 0.81 \\
\bottomrule
\end{tabular}
\caption{
Unit-level agreement on whether each candidate intent or
detail unit is present in the query. Computed on a candidate-set pool drawn from 50 queries.
}
\label{tab:human_exp2}
\end{table}

\subsection{Experiment 3: Does the LLM Extractor Recover the Same Information Units as Humans?}
\label{sec:appendix_exp3}

\noindent\textbf{Setup.}
The LLM extractor in our evaluation pipeline maps a simulated-user reply
to the set of information units the reply discloses. To check that this
extractor agrees with human judgement, we sampled $50$ dialogue turns from
completed VagueBench evaluations, each consisting of the assistant's
clarifying question, the simulated user's reply, and the set of candidate
units that could plausibly be disclosed at that turn. For every candidate
unit, each annotator independently judged Disclosed or Not Disclosed by
the reply, and the LLM extractor contributed its own Disclosed or Not
Disclosed label on the same candidate set. Agreement is again computed at
the unit level, pooled over all $50$ turns.

\noindent\textbf{Results.}
Table~\ref{tab:human_exp3} reports the three pairwise Cohen's $\kappa$
values. Inter-annotator $\kappa$ was $0.84$, comparable to Experiment~2.
The LLM extractor agreed with Human~1 at $\kappa = 0.80$ and with
Human~2 at $\kappa = 0.79$. The closeness of these three numbers
indicates that the extractor is interchangeable with a human annotator
for the purpose of unit-level recovery, and therefore that Cov.~F1 and
IAR as measured against the extractor's output are reliable proxies for
what a human would have scored.

\begin{table}[htbp]
\centering
\small
\setlength{\tabcolsep}{8pt}
\renewcommand{\arraystretch}{1.15}
\begin{tabular}{l c}
\toprule
\textbf{Pair} & Cohen's $\kappa$ \\
\midrule
Human~1 vs Human~2  & 0.84 \\
Human~1 vs LLM      & 0.80 \\
Human~2 vs LLM      & 0.79 \\
\bottomrule
\end{tabular}
\caption{
Unit-level agreement on whether each candidate unit was
disclosed by a simulated-user reply. Computed on a candidate-set pool
drawn from 50 dialogue turns.
}
\label{tab:human_exp3}
\end{table}

\section{Simulated User}
\label{sec:appendix_simulated_user}
 
This appendix specifies the simulated user $\mathcal{S}$ \citep{shi2019build, hong2023zero,faltings2023interactive, hu2023unlocking}introduced in \S\ref{sec:evaluation}. $\mathcal{S}$ is the core component of the VagueBench evaluation protocol: it plays the role of a real user who holds the hidden information and decides, on each turn, which units to release in response to the model's question. Two properties anchor its design: \emph{consistency}---identical questions elicit responses of the same kind---and \emph{minimal disclosure}---only precisely-matched units are released, with no volunteered extras.

\subsection{Behaviour Rules}
\label{sec:appendix_simulator_rules}

The behaviour of $\mathcal{S}$ is governed by five rules over the hidden intent set $\mathcal{T}_{\text{hidden}}$ and detail set $\mathcal{D}_{\text{hidden}}$.

\paragraph{R1. Information release.}
When the model's question semantically matches the trigger keywords of some hidden intent $I_j \in \mathcal{T}_{\text{hidden}}$ or detail $d_i \in \mathcal{D}_{\text{hidden}}$, $\mathcal{S}$ releases that unit in short, natural language. For example, the question \emph{``Which area would you like to dine in?''} matches the trigger of $\langle\texttt{area},\;\text{near University Avenue}\rangle$ and $\mathcal{S}$ replies \emph{``near University Avenue in downtown Palo Alto''}; an intent-side match is symmetric, e.g.\ \emph{``Are you looking for restaurant recommendations or something else?''} matches the \texttt{recommend\_restaurant} intent and $\mathcal{S}$ replies \emph{``yes, recommend a few restaurants''}. Matching is judged at the semantic level by a strong LLM rather than by exact keyword overlap, so phrasings such as \emph{``roughly how much do you want to spend?''} still match a budget trigger listing \emph{budget~/~price~/~spending}.

\paragraph{R2. Deflection.}
If the question matches no hidden unit, $\mathcal{S}$ returns a deflective reply drawn at random from a predefined pool: \emph{``I haven't decided yet''}, \emph{``I'm not sure about that''}, \emph{``you decide for me''}, \emph{``no particular preference''}, and similar. Deflections are short, natural, and carry no cues that the model could use to infer the hidden set.

\paragraph{R3. No proactive disclosure.}
$\mathcal{S}$ never volunteers a unit that was not asked about. Questions that approach several units but precisely match only one---for example \emph{``any requirements for your meal?''} matching only \texttt{scenario}---release only the matched unit, regardless of how close the others appear.

\paragraph{R4. Release once.}
Once released, a unit is removed from $\mathcal{T}_{\text{hidden}}$ or $\mathcal{D}_{\text{hidden}}$. Subsequent questions that re-hit the same unit are answered with \emph{``I mentioned that earlier''} or by repeating the previous value; no new credit is granted at evaluation.

\paragraph{R5. Multi-unit matches.}
When a single question matches multiple hidden units---for example \emph{``which city and what occasion?''} hitting both \texttt{city} and \texttt{scenario}---$\mathcal{S}$ releases all matched units in one reply. This rule ensures that elicitation efficiency is not artificially deflated by simulator-side fragmentation.

\subsection{Direct-Answer Behaviour}
\label{sec:appendix_simulator_no_ask}

If the model commits to a final answer at the initial turn without asking, the dialogue terminates immediately and the evaluation proceeds without further simulator input. On information-complete queries this is the expected behaviour and requires no simulator participation. On Intent-Vague and Fully-Vague queries, the absence of a question means the model answered under missing information, and the resulting answer is naturally penalised by Cov.~F1, which compares the response against the gold information set. Whether a model turn counts as \emph{ask} or \emph{answer} is judged by the LLM-based behaviour judge (Prompt~3 in Appendix~\ref{sec:appendix_prompts}).

\subsection{System-Prompt Template}
\label{sec:appendix_simulator_prompt}

Each Intent-Vague and Fully-Vague sample carries a system prompt that instantiates $\mathcal{S}$ for that sample. The template has four components:
\begin{itemize}[leftmargin=*, itemsep=2pt, parsep=0pt]
    \item \textbf{Role description.} The complete underlying need together with the meta query actually shown to the model.
    \item \textbf{Expressed information.} The content already present in the meta query, which $\mathcal{S}$ does not need to release.
    \item \textbf{Hidden-information table.} A structured listing of every hidden intent and detail, each annotated with its trigger keywords and the natural-language reply to use on release.
    \item \textbf{Deflection pool.} The set of non-informative replies drawn upon when no unit is matched.
\end{itemize}

A representative Fully-Vague system prompt is shown in Figure~\ref{fig:simulator_prompt_example}.

\begin{figure}[t]
\begin{tcolorbox}[
    colback=gray!5!white,
    colframe=gray!75!black,
    fontupper=\footnotesize,
    left=4pt, right=4pt, top=4pt, bottom=4pt, boxsep=2pt,
]
\textbf{Role.} You are a user asking an AI assistant for help. Your actual need is: a recommendation for a Japanese restaurant near University Avenue in downtown Palo Alto suitable for business dining, around \$120 per person, ideally with a private room. You only said: \emph{``Palo Alto Japanese food''}.

\vspace{0.5mm}
\textbf{Hidden intents.}
\begin{itemize}[leftmargin=*, itemsep=0pt, parsep=0pt, label=$\bullet$]
    \item \texttt{recommend\_restaurant} --- if the AI asks what you want / what kind of help / your goal $\rightarrow$ reply \emph{``recommend a few restaurants''}.
\end{itemize}

\textbf{Hidden details.}
\begin{itemize}[leftmargin=*, itemsep=0pt, parsep=0pt, label=$\bullet$]
    \item \texttt{area} = near University Avenue --- if the AI asks about location / district / where $\rightarrow$ reply \emph{``near University Avenue in downtown Palo Alto''}.
    \item \texttt{scenario} = business dining --- if the AI asks about occasion / company / purpose $\rightarrow$ reply \emph{``business dining''}.
    \item \texttt{budget} = around \$120 per person --- if the AI asks about budget / price / spending $\rightarrow$ reply \emph{``around \$120 per person''}.
    \item \texttt{private\_room} = required --- if the AI asks about ambience / seating / private rooms $\rightarrow$ reply \emph{``private room preferred''}.
\end{itemize}

\textbf{Deflection.} For unmatched questions, randomly reply with one of \emph{``I don't have particular preferences''}, \emph{``you decide for me''}, \emph{``I haven't thought about that''}.

\vspace{0.5mm}
\textbf{Notes.} Reply briefly and naturally, like a real user. Do not volunteer information that was not asked about. Answer only what was asked.
\end{tcolorbox}
\caption{\textbf{System prompt example} for a Fully-Vague sample. The hidden-information table lists the four detail units and one intent unit that $\mathcal{S}$ may release on a matching question.}
\label{fig:simulator_prompt_example}
\end{figure}

\subsection{Implementation}

$\mathcal{S}$ is realised with GLM-5.1, which receives the per-sample system prompt and produces a reply to each model question. Semantic matching between the question and the hidden-unit triggers is performed inside the same inference call: the LLM judges, on the basis of the trigger keywords and the hidden-information table in the system prompt, whether the model's question hits any unit, and emits a release reply or a deflection accordingly. To keep simulator behaviour consistent and auditable, we set the sampling temperature to $0.1$ to suppress drift and log every simulator turn for post-hoc inspection and spot-checking.

\section{Implementation Details}
\label{sec:appendix_implementation}

This appendix lists the models evaluated in the experiments of \S\ref{sec:experiments}, the hyperparameters used to train DialogGRPO, and the hardware on which training and inference were carried out.

\paragraph{Evaluated models.}
Table~\ref{tab:appendix_models} enumerates the closed and open-source models compared on VagueBench and the two transfer benchmarks. Closed-source models are accessed via their official APIs; open-source models are loaded from the Hugging Face checkpoints linked in the table.

\begin{table}[t!]
\centering
\small
\renewcommand{\arraystretch}{1.1}
\begin{tabularx}{\columnwidth}{@{}l>{\raggedright\arraybackslash}Xl@{}}
\toprule
\textbf{Provider} & \textbf{Model} & \textbf{Model Link} \\
\midrule
\multirow{2}{*}{OpenAI}      & GPT-5.4               & \href{https://platform.openai.com/docs/models}{OpenAI Docs} \\
 & GPT-oss-120B & \href{https://huggingface.co/openai/gpt-oss-120b}{Hugging Face} \\
\midrule
Anthropic                    & Claude Sonnet 4.6      & \href{https://www.anthropic.com/claude}{Anthropic} \\
\midrule
\multirow{2}{*}{Google}
                      & Gemini 3.1 Pro         & \href{https://deepmind.google/technologies/gemini/}{Google DeepMind} \\
                       & Gemma3-27B         & \href{https://huggingface.co/google/gemma-3-27b-it}{Hugging Face} \\
\midrule
Meta                         & Llama-3.1-8B-Instruct  & \href{https://huggingface.co/meta-llama/Llama-3.1-8B-Instruct}{Hugging Face} \\
\midrule
\multirow{2}{*}{Qwen}        & Qwen3-8B               & \href{https://huggingface.co/Qwen/Qwen3-8B}{Hugging Face} \\
                             & Qwen3.5-27B            & \href{https://huggingface.co/Qwen/Qwen3.5-27B}{Hugging Face} \\
\midrule
\multirow{2}{*}{Zhipu}       & GLM-5                  & \href{https://huggingface.co/zai-org/GLM-5}{Hugging Face} \\
                             & GLM-5.1   & \href{https://huggingface.co/zai-org/GLM-5.1}{Hugging Face} \\
\bottomrule
\end{tabularx}
\caption{Models evaluated in this work. The same Qwen3.5-27B checkpoint serves as a base-model baseline (Table~\ref{tab:main}); the cross-base experiment in \S\ref{sec:generalization} additionally trains Gemma3-27B with identical hyperparameters. GLM-5.1 plays the simulated user $\mathcal{S}$ (Appendix~\ref{sec:appendix_simulated_user}).}
\label{tab:appendix_models}
\end{table}

\paragraph{Training hyperparameters.}
DialogGRPO is trained with LoRA adapters on top of the Qwen3.5-27B base; the Gemma3-27B run in \S\ref{sec:generalization} uses the same configuration with only the base weights swapped. Table~\ref{tab:appendix_hyperparams} reports the full set of training hyperparameters.

\begin{table}[t]
\centering
\small
\setlength{\tabcolsep}{6pt}
\renewcommand{\arraystretch}{1.15}
\begin{tabular}{@{}ll@{}}
\toprule
\textbf{Hyperparameter} & \textbf{Value} \\
\midrule
Learning rate          & $10^{-4}$ \\
Warm-up ratio          & $0.1$ \\
LR scheduler           & cosine \\
Batch size             & $48$ \\
Epochs                 & $3$ \\
LoRA rank              & $16$ \\
LoRA alpha             & $32$ \\
\bottomrule
\end{tabular}
\caption{Training hyperparameters for DialogGRPO. The same configuration is used in the cross-base-model experiment of \S\ref{sec:generalization}.}
\label{tab:appendix_hyperparams}
\end{table}

\paragraph{Hardware.}
All training and inference runs are carried out on a single node equipped with eight NVIDIA~H200 GPUs. DialogGRPO training of Qwen3.5-27B and the cross-base Gemma3-27B run in \S\ref{sec:generalization} both use the same eight-GPU node; open-source baseline inference and the simulated-user rollouts used for evaluation are also run on this hardware.

\section{General Capability Preservation}
\label{sec:appendix_general_capability}

A natural concern about any offline reinforcement learning recipe is that
the trained model may have drifted away from the base distribution and
lost general task ability. To address this, we evaluated our DialogGRPO
model on six general-capability benchmarks alongside the un-modified
Qwen3.5-27B base, covering coding, mathematics, commonsense, and broad
knowledge: LiveCodeBench \citep{livecodebench}, MathQA \citep{mathqa},
CommonsenseQA \citep{csqa}, MATH-500 \citep{math}, MMLU \citep{mmlu}, and
GSM8K \citep{gsm8k}. No clarification-related training data overlap with
any of these benchmarks.

\begin{figure}[htbp]
\centering
\includegraphics[width=0.95\linewidth]{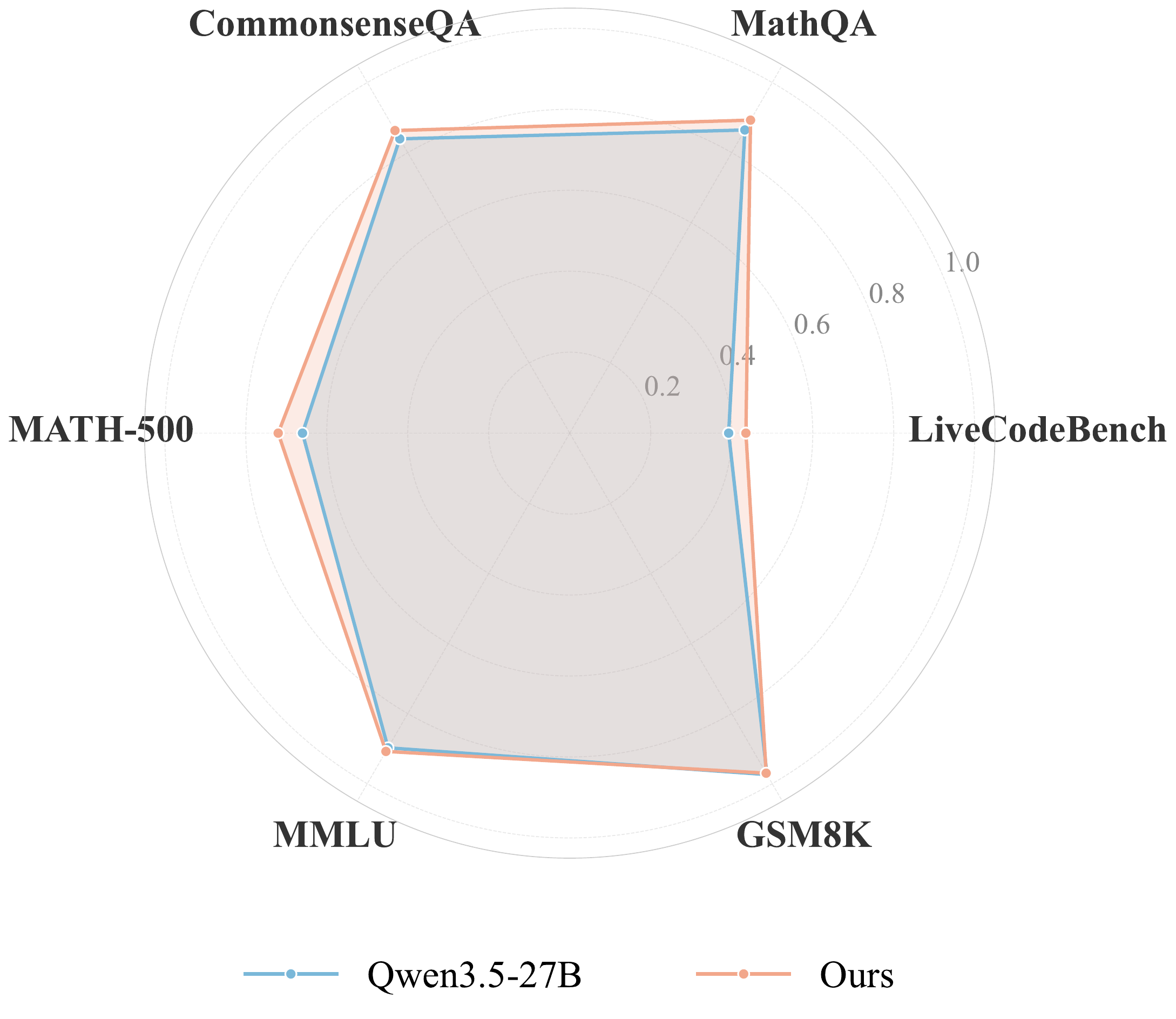}
\caption{
General-capability check on six external benchmarks: LiveCodeBench,
MathQA, CommonsenseQA, MATH-500, MMLU, and GSM8K. The two envelopes trace
nearly identical shapes, with the trained DialogGRPO model staying at or
above the Qwen3.5-27B base on every benchmark and slightly improving on
LiveCodeBench and MATH-500. The result indicates that the proactive
elicitation behaviour reported in the main text is acquired without
sacrificing general task ability.
}
\label{fig:radar_general}
\end{figure}

Figure~\ref{fig:radar_general} shows the comparison as a radar chart. The
two envelopes are nearly identical, with our model matching the base
within measurement noise on four of the six benchmarks and slightly
exceeding it on LiveCodeBench and MATH-500. The result mitigates a
potential reviewer concern: the proactive elicitation behaviour reported
in the main text is acquired without sacrificing general task accuracy.
We attribute this to the nature of what DialogGRPO trains. The objective
optimises \emph{when} and \emph{what} to ask, a decision-making behaviour
over the dialogue state, rather than injecting domain knowledge into the
model; it reshapes how the model acts on missing information without
overwriting the factual, mathematical, or coding competence already
present in the base weights. This behaviour-level target is reinforced by
the diversity of our training data, which spans five task families and a
broad range of open-domain topics rather than any single narrow domain.
Because no benchmark-specific knowledge is added and no narrow
distribution is over-fit, the elicitation policy is learned on top of the
base model's general ability rather than at its expense.

\section{Generalization to a Different Base Model}
\label{sec:appendix_gemma}

This appendix expands on the cross-base-model generalization experiment summarised in \S\ref{sec:generalization} and Figure~\ref{fig:cross_base}.

\paragraph{Setup.}
To check that our method generalises beyond a single base model, we re-ran the full training pipeline on Gemma3-27B, keeping every hyperparameter identical to the Qwen3.5-27B run reported in the main text. The DAG construction, reward design were unchanged; only the base-model weights differed.

\paragraph{The gains carry over.}
Gemma3-27B shows the same suppression pattern as Qwen3.5-27B before training, with the base Ask\% below $3\%$ on both splits, confirming that clarification suppression is shared across base-model families rather than being a property of one model. After training the Gemma version reaches Ask\%, Cov.~F1, and IAR levels that closely track those observed on Qwen3.5-27B, and the lift bars in Figure~\ref{fig:cross_base} are of similar height on every metric. The consistency across two unrelated base models indicates that the training recipe transfers, and that the improvements in the main paper are attributable to the DialogGRPO  design rather than to a peculiarity of any single base model.

\section{Additional Experiments}
\label{sec:appendix_additional_experiments}

We additionally compare DialogGRPO with CollabLLM~\citep{wu2025collabllm} and test transfer to a sub-10B backbone. Because the released CollabLLM checkpoint uses Llama-3.1-8B-Instruct, we re-ran the full DialogGRPO pipeline on the same backbone for a controlled same-backbone comparison.

\begin{table*}[t]
\centering
\small
\setlength{\tabcolsep}{9pt}
\renewcommand{\arraystretch}{1.1}
\begin{tabular}{l ccc ccc}
\toprule
\multirow{2}{*}{\textbf{Method (Llama-3.1-8B-Instruct)}}
& \multicolumn{3}{c}{\textbf{Intent-Vague}}
& \multicolumn{3}{c}{\textbf{Fully-Vague}} \\
\cmidrule(lr){2-4} \cmidrule(lr){5-7}
& Ask\% & Cov. F1 & IAR & Ask\% & Cov. F1 & IAR \\
\midrule
Base                & 0.8  & 0.50 & 0.00 & 1.2  & 0.35 & 0.23 \\
CollabLLM           & 3.2  & 0.59 & 0.00 & 4.4  & 0.37 & 0.27 \\
\rowcolor{gray!10}
\textbf{DialogGRPO} & \textbf{76.6} & \textbf{0.61} & \textbf{0.37}
                    & \textbf{82.0} & \textbf{0.49} & \textbf{0.43} \\
\bottomrule
\end{tabular}
\caption{Same-backbone comparison with CollabLLM and transfer of DialogGRPO to a sub-10B model on VagueBench.}
\label{tab:appendix_llama_collabllm}
\end{table*}

CollabLLM produces only modest gains over the base model on VagueBench, increasing Ask\% to $3.2$/$4.4$. This is consistent with its focus on vertical tasks such as document editing, code, and mathematics, using task-specific rewards and online user simulation rather than open-domain recovery of never-stated information. On the same Llama backbone, DialogGRPO reaches $76.6$/$82.0$ Ask\% while improving both Cov.~F1 and IAR. The suppression-then-recovery pattern therefore also holds at the 8B scale, providing evidence that the training recipe transfers beyond the two 27B backbones reported in \S\ref{sec:generalization}.

\section{Derivation of the Advantage-Weighted Objective}
\label{sec:appendix_derivation}

This appendix derives the training objective in
Eq.~\ref{eq:objective} from a KL-regularized policy optimization
problem. The derivation follows the classical reward-weighted
regression line \citep{rwr, peng2019advantage}; the same closed-form
step underlies the optimal policy in DPO \citep{dpo}.
We make explicit how the KL constraint is absorbed into an exponential
advantage weight, which is why Eq.~\ref{eq:objective} carries no
separate KL term.

\paragraph{KL-regularized objective.}
For a state $s$ (a canonical prompt $p_v$) with candidate actions $a$
(responses $y_e$), we seek a policy $\pi$ that maximizes the advantage
$A(s,a)=\hat{A}_e$ while staying close to the reference policy
$\pi_{\mathrm{ref}}$:
\begin{equation}
\begin{split}
\max_{\pi}\ \mathbb{E}_{s\sim\mathcal{D}}
\bigl[ & \sum_a \pi(a\mid s)\,A(s,a) \\
& - \beta\,\mathrm{KL}\!\left(\pi(\cdot\mid s)\,\|\,\pi_{\mathrm{ref}}(\cdot\mid s)\right) \bigr] ,
\end{split}
\label{eq:app_klobj}
\end{equation}
subject to the normalization constraint
$\sum_a \pi(a\mid s)=1$ for every $s$.

\paragraph{Closed-form optimum.}
Introducing a Lagrange multiplier $\lambda(s)$ for the normalization
constraint at each state and setting the derivative of the Lagrangian
with respect to $\pi(a\mid s)$ to zero gives
\begin{equation}
A(s,a) - \beta\log\frac{\pi(a\mid s)}{\pi_{\mathrm{ref}}(a\mid s)}
- \beta - \frac{\lambda(s)}{P(s)} = 0 .
\end{equation}
Solving for $\pi$ and folding all $a$-independent terms into a
per-state normalizer $Z(s)$ yields the optimal policy
\begin{equation}
\pi^{\star}(a\mid s)
= \frac{1}{Z(s)}\,\pi_{\mathrm{ref}}(a\mid s)\,
\exp\!\Big(\tfrac{1}{\beta}A(s,a)\Big) ,
\label{eq:app_optpolicy}
\end{equation}
where $Z(s)=\sum_a \pi_{\mathrm{ref}}(a\mid s)\exp(A(s,a)/\beta)$
ensures $\sum_a\pi^{\star}(a\mid s)=1$.
The KL constraint in Eq.~\ref{eq:app_klobj} is thus not discarded: it is
absorbed analytically into the exponential factor $\exp(A/\beta)$ and
the reference term $\pi_{\mathrm{ref}}$.

\paragraph{Projection onto a parametric policy.}
$Z(s)$ requires summing over all actions and is intractable, so
$\pi^{\star}$ cannot be used directly. We instead fit a parametric
policy $\pi_\theta$ to $\pi^{\star}$ by minimizing the forward KL
divergence $\mathrm{KL}(\pi^{\star}\,\|\,\pi_\theta)$, which, after
dropping $\theta$-independent terms, is equivalent to
\begin{equation}
\max_{\theta}\ \mathbb{E}_{s}
\left[\sum_a \pi^{\star}(a\mid s)\,\log\pi_\theta(a\mid s)\right].
\end{equation}
Substituting Eq.~\ref{eq:app_optpolicy} gives
\begin{equation}
\begin{split}
\max_{\theta}\ \mathbb{E}_{s}
\bigg[ \sum_a & \tfrac{1}{Z(s)}\,
\pi_{\mathrm{ref}}(a\mid s) \\
& \cdot \exp\!\Big(\tfrac{A(s,a)}{\beta}\Big)\,
\log\pi_\theta(a\mid s) \bigg] .
\end{split}
\end{equation}

\paragraph{Empirical objective.}
In our offline setting the candidate set $\mathcal{G}_v$ at each node is
pre-generated by the external strong LLM, so it does not coincide with
samples from $\pi_{\mathrm{ref}}$. We therefore treat the candidates as
an empirical sampling distribution and approximate the expectation over
$\pi_{\mathrm{ref}}(a\mid s)$ by an average over $\mathcal{G}_v$; this is
a stated approximation rather than an exact identity, and is made
practical by the advantage weighting, which down-weights candidates far
from high-advantage regions. Since $Z(s)$ is independent of $\theta$ and
acts only as a per-state scaling, it can be dropped during optimization.
This yields the per-node objective in Eq.~\ref{eq:objective}.
Unlike trajectory-level reward-weighted regression
\citep{rwr}, the advantage $\hat{A}_e$ here is computed per
edge within a dialogue-state DAG (\S\ref{sec:turn_reward}), giving dense
turn-level credit assignment rather than a single trajectory return.

\section{Case Studies}
\label{sec:appendix_case_study}

This appendix shows three trajectories produced by DialogGRPO on
representative test queries, rendered as chat transcripts. Each turn
shows a clarifying question asked by the model on the left and the
simulated user's reply on the right; the green panel at the bottom
summarises the final answer with its key points highlighted.

\newcommand{\botbubble}[1]{%
  \begin{tcolorbox}[colback=white, colframe=gray!35, boxrule=0.3pt,
    arc=2.5mm, boxsep=1pt, left=6pt, right=6pt, top=2pt, bottom=2pt,
    fontupper=\small]#1\end{tcolorbox}}
\newcommand{\userbubble}[1]{%
  \begin{tcolorbox}[colback=gray!8, colframe=gray!35, boxrule=0.3pt,
    arc=2.5mm, boxsep=1pt, left=6pt, right=6pt, top=2pt, bottom=2pt,
    fontupper=\small]#1\end{tcolorbox}}
\newcommand{\thinkline}[1]{%
  {\small\itshape\color{gray!55!black}%
   \texttt{<think>}~\ldots~#1~\texttt{</think>}}}
\newcommand{\botturn}[3]{
  \par\noindent
  \makebox[0.09\linewidth][l]{\small\textbf{Turn~#1}}%
  \begin{minipage}[t]{0.70\linewidth}%
    \thinkline{#2}\par\vspace{1pt}%
    \botbubble{#3}%
  \end{minipage}\par\vspace{3pt}}
\newcommand{\userturn}[1]{%
  \par\noindent\hspace*{\fill}%
  \begin{minipage}[t]{0.70\linewidth}\raggedleft\userbubble{#1}\end{minipage}%
  \par\vspace{3pt}}
\newcommand{\answerthink}[1]{%
  \par\noindent\hspace*{0.09\linewidth}%
  \begin{minipage}[t]{0.82\linewidth}\thinkline{#1}\end{minipage}%
  \par\vspace{2pt}}
\newcommand{\chk}{\textcolor{green!55!black}{\ding{51}}}

\begin{figure*}[!t]
\centering
\begin{tcolorbox}[
    enhanced,
    colback=white, colframe=white, boxrule=0pt,
    borderline={0.7pt}{0pt}{teal!55!black, dashed},
    arc=3mm,
    left=10pt, right=10pt, top=18pt, bottom=10pt,
    title={\textbf{Proactive Elicitation}},
    coltitle=teal!55!black,
    fonttitle=\small\sffamily,
    attach boxed title to top left={xshift=12pt, yshift=-7pt},
    boxed title style={colback=white, colframe=white, boxrule=0pt,
        size=small, top=1pt, bottom=1pt, left=4pt, right=4pt},
]
\userturn{Ibuprofen and acetaminophen.}
\botturn{1}{Two drugs named, but the user gave no \textbf{intent} and no \textbf{symptom}; cannot assume the use case $\Rightarrow$ \textbf{ask}}{What would you like to know --- their differences, or whether they can be taken together?}
\userturn{The difference, which is better, whether to alternate, and how long to wait between doses.}
\botturn{2}{Intent clear; the \textbf{symptom} and the \textbf{target user} are still missing, both change the recommendation $\Rightarrow$ \textbf{ask}}{What symptom are you trying to relieve --- fever, headache, menstrual pain --- and who is taking it?}
\userturn{Just a cold with fever; I'm taking it myself.}

\answerthink{All key units acquired $\Rightarrow$ \textbf{answer}}
\begin{center}
\begin{tcolorbox}[colback=green!8, colframe=green!55!black,
    boxrule=0.5pt, arc=2mm, boxsep=1pt,
    left=8pt, right=8pt, top=3pt, bottom=3pt,
    width=0.82\linewidth, fontupper=\small]
\chk~\textbf{Acetaminophen}: gentler, safer; weaker anti-inflammatory.\\
\chk~\textbf{Ibuprofen}: stronger fever and pain relief; harder on the stomach.\\
\chk~\textbf{Alternating: not recommended} for self-medication.\\
\chk~Intervals: ibuprofen \textbf{6--8\,h}, acetaminophen \textbf{4--6\,h}; max 4 doses per day.
\end{tcolorbox}
\end{center}
\end{tcolorbox}
\caption{\textbf{Case 1.} Drug comparison --- the model elicits the user's intent, then the symptom and the target user, before delivering a focused recommendation.}
\label{fig:case1}
\end{figure*}

\begin{figure*}[!t]
\centering
\begin{tcolorbox}[
    enhanced,
    colback=white, colframe=white, boxrule=0pt,
    borderline={0.7pt}{0pt}{teal!55!black, dashed},
    arc=3mm,
    left=10pt, right=10pt, top=18pt, bottom=10pt,
    title={\textbf{Proactive Elicitation}},
    coltitle=teal!55!black,
    fonttitle=\small\sffamily,
    attach boxed title to top left={xshift=12pt, yshift=-7pt},
    boxed title style={colback=white, colframe=white, boxrule=0pt,
        size=small, top=1pt, bottom=1pt, left=4pt, right=4pt},
]
\userturn{Short-term stocks for next Monday.}
\botturn{1}{Time, strategy, and asset type are stated; the \textbf{specific intent}, recommendation versus analysis, is missing $\Rightarrow$ \textbf{ask}}{Are you asking for short-term recommendations for next Monday, or technical analysis of a particular sector or stock?}
\userturn{Short-term recommendations --- hot sectors or direct tickers.}
\botturn{2}{Hot short-term plays swing widely; the user's \textbf{risk preference} is not stated $\Rightarrow$ \textbf{ask}}{Hot short-term plays swing widely. Aggressive moves such as chasing limit-up stocks, or something more conservative?}
\userturn{Aggressive is fine; short-term is about chasing returns. Higher risk is OK.}
\botturn{3}{Risk profile clear; the \textbf{investor identity}, retail vs.\ institutional, still affects liquidity matching $\Rightarrow$ \textbf{ask}}{Are you an individual retail investor? Liquidity needs differ between retail and institutional accounts.}
\userturn{Yes, retail.}

\answerthink{All key units acquired $\Rightarrow$ \textbf{answer}}
\begin{center}
\begin{tcolorbox}[colback=green!8, colframe=green!55!black,
    boxrule=0.5pt, arc=2mm, boxsep=1pt,
    left=8pt, right=8pt, top=3pt, bottom=3pt,
    width=0.82\linewidth, fontupper=\small]
\chk~Focus on \textbf{leading names in currently hot sectors}, e.g.\ technology or new energy.\\
\chk~Track \textbf{capital flow at Monday's open}; enter the strongest names on confirmation.\\
\chk~Always preset a \textbf{stop-loss} given the volatility.\\
\chk~Compliance limits prevent listing tickers; pair sector analysis with a licensed advisor.
\end{tcolorbox}
\end{center}
\end{tcolorbox}
\caption{\textbf{Case 2.} Short-term stock query --- the model successively elicits intent, risk preference, and investor identity before delivering a strategy sketch.}
\label{fig:case2}
\end{figure*}

\begin{figure*}[!t]
\centering
\begin{tcolorbox}[
    enhanced,
    colback=white, colframe=white, boxrule=0pt,
    borderline={0.7pt}{0pt}{teal!55!black, dashed},
    arc=3mm,
    left=10pt, right=10pt, top=18pt, bottom=10pt,
    title={\textbf{Proactive Elicitation}},
    coltitle=teal!55!black,
    fonttitle=\small\sffamily,
    attach boxed title to top left={xshift=12pt, yshift=-7pt},
    boxed title style={colback=white, colframe=white, boxrule=0pt,
        size=small, top=1pt, bottom=1pt, left=4pt, right=4pt},
]
\userturn{RPG bounty-quest reward design.}
\botturn{1}{Game type and target system stated; the \textbf{specific intent} --- new ideas, code, or balancing --- is missing $\Rightarrow$ \textbf{ask}}{What specifically are you after --- innovative design directions, or help with a concrete issue in your current setup?}
\userturn{Innovative design ideas.}
\botturn{2}{Intent is innovation; the \textbf{current mechanism} and the \textbf{pain point} are unknown, otherwise advice will be generic $\Rightarrow$ \textbf{ask}}{Could you describe the current reward mechanism, or what feels most lacking in the existing design?}
\userturn{Stale: kill the target, get gold, XP, maybe a random drop. No surprise, no incentive to chase hard bounties.}

\answerthink{Current state and pain point clear $\Rightarrow$ \textbf{answer}}
\begin{center}
\begin{tcolorbox}[colback=green!8, colframe=green!55!black,
    boxrule=0.5pt, arc=2mm, boxsep=1pt,
    left=8pt, right=8pt, top=3pt, bottom=3pt,
    width=0.82\linewidth, fontupper=\small]
\chk~\textbf{Dynamic bounty pool}: reward scales with evasion time and failed attempts.\\
\chk~\textbf{Bounty tokens and self-chosen rewards}: random drops become a goal-directed grind.\\
\chk~\textbf{Non-numerical rewards and identity}: exclusive titles, statues, special shops, leaderboard.\\
\chk~\textbf{Random affixes and mystery chests}: roguelike modifiers and rare-jackpot chests.
\end{tcolorbox}
\end{center}
\end{tcolorbox}
\caption{\textbf{Case 3.} RPG reward design --- the model elicits intent, then the current mechanism and pain point, before delivering targeted design ideas.}
\label{fig:case3}
\end{figure*}
\section{Prompt Templates}
\label{sec:appendix_prompts}

This section collects the four prompt templates that drive VagueBench's
construction and evaluation pipelines. Variables enclosed in
\texttt{\{ \}} are filled with run-time values. The four prompts map to
the paper as follows: Prompt~1 is the seed filter referenced in
\S\ref{sec:data_collection}; Prompt~2 is the variant-synthesis prompt
described in \S\ref{sec:data_construction}; Prompt~3 is the behavior
judge that produces the Ask\% metric in \S\ref{sec:evaluation}; Prompt~4
is the information extractor that produces Cov.~F1 and supports IAR in
the same section.

For the model evaluation, we additionally use two system prompts. In the
naive setting, the system prompt is ``\texttt{You are a helpful
assistant.}'' In the question-guided setting, we use ``\texttt{You are a
helpful assistant. When you think it is necessary, you may ask the user a
question. Ask at most one question at a time.}''

\begin{figure*}[!t]
\begin{tcolorbox}[
    colback=gray!5!white,
    colframe=gray!75!black,
]
\textbf{[System Prompt]} \\
Determine whether the given query qualifies as an information-complete
\emph{seed query} for VagueBench.

A seed query must simultaneously satisfy all of the following conditions:
\begin{enumerate}[leftmargin=*, itemsep=2pt, parsep=0pt]
    \item \textbf{Clear intent.} The query contains explicit intent
          anchors: precise action verbs, explicit speech-act framing,
          or clear output expectations, so that the user's task can
          be identified directly.
    \item \textbf{Sufficient details.} The query carries the key
          parameters and constraints needed to execute that task, so
          that a competent assistant could produce a high-quality answer
          without asking any clarifying question.
    \item \textbf{Independent and self-contained.} The query does not
          depend on images, attached files, or prior dialogue turns; it
          can be understood out of context.
    \item \textbf{Compliant content.} The query is not pure chit-chat,
          a test string such as ``hello'' or ``test'', policy-violating
          content, or a code-debugging snippet pasted with no
          accompanying question.
\end{enumerate}

\textbf{Reject the query} if any of the following holds:
\begin{itemize}[leftmargin=*, itemsep=2pt, parsep=0pt]
    \item \textit{Vague intent.} Only entities or keywords without an
          action intent, for example ``Tesla'' or ``Shanghai
          restaurants''.
    \item \textit{Missing details.} The intent is clear but key
          parameters are absent, for example ``Help me write an email''
          without recipient, subject, or content.
    \item \textit{Context-dependent.} The query uses pronouns or
          references such as ``what I just said'', ``the one above'',
          or ``help me modify''.
    \item \textit{Privacy.} The query exposes the user's personal
          private information.
\end{itemize}

Output JSON: \texttt{\{"is\_seed": true|false, "reason": "One-sentence reason"\}}
\end{tcolorbox}
\caption{\textbf{Prompt 1: Seed Query Filter.}}
\label{fig:prompt1}
\end{figure*}

\begin{figure*}[!t]
\begin{tcolorbox}[
    colback=gray!5!white,
    colframe=gray!75!black,
    fontupper=\scriptsize,
    left=3pt, right=3pt, top=3pt, bottom=3pt,
    boxsep=1pt,
]
\textbf{[System Prompt]} \\
You are a dialogue-data annotation expert. Given an
information-complete seed query, annotate it into the information set
$\mathcal{I} = (\mathcal{T}, \mathcal{D})$ and synthesise the
\textbf{Intent-Vague} and \textbf{Fully-Vague} variants following the
controlled-degradation procedure.

\vspace{0.5mm}
\textbf{\#\# Step 1: Query-type classification.} \\
Choose exactly one type from: Information Query / Recommendation \&
Decision / Task Execution / Creation \& Generation / Consultation \&
Analysis.

\vspace{0.5mm}
\textbf{\#\# Step 2: Annotate
\boldmath$\mathcal{I} = (\mathcal{T},\mathcal{D})$\unboldmath.} \\
\textbf{Intent set $\mathcal{T}$.} All task-level goals the user expects
to complete, each represented as $\{id, label\}$. Examples of intent
labels: ``recommend restaurant'', ``translate text'', ``compare plans''.
A seed may have a single intent or multiple parallel intents.

\textbf{Detail set $\mathcal{D}$.} Specific parameters and constraints
required to execute the intents, each represented as
$\{id, key, value, intent\_ids\}$. Examples of detail keys: city, budget,
scenario, cuisine. Each detail records the list of intents it serves.

\vspace{0.5mm}
\textbf{\#\# Step 3: Synthesise the Intent-Vague variant.} \\
\textbf{Goal.} Weaken or remove \emph{only} the intent-layer anchors
while preserving every detail-layer anchor and every detail unit in
$\mathcal{D}$. After rewriting it should
be impossible to identify the user's intended task, yet every
parameter, constraint, output-type cue, and scope cue must remain.

\textbf{Operations.}
\begin{itemize}[leftmargin=*, itemsep=2pt, parsep=0pt]
    \item \textit{Predicate anchors} (precise action verbs)
          $\rightarrow$ \emph{Generalise}.
          For example: ``summarize'' $\rightarrow$ ``work on'',
          ``translate'' $\rightarrow$ ``process'', or drop the verb
          entirely.
    \item \textit{Speech-act anchors} (sentence frames such as
          ``Help me X'', ``I need X'') $\rightarrow$ \emph{Weaken or
          remove}. For example: ``Help me X'' $\rightarrow$ ``About X'';
          ``I need X'' $\rightarrow$ ``How is X''.
    \item \textit{Output-type anchors} and \textit{scope anchors}
          (detail layer) $\rightarrow$ \emph{Keep verbatim}.
\end{itemize}

\textbf{Rewriting rules.}
\begin{itemize}[leftmargin=*, itemsep=2pt, parsep=0pt]
    \item Every detail-layer expression and every detail unit in
          $\mathcal{D}$ must be retained.
    \item The rewritten text must be natural and fluent, resembling a
          real underspecified user message.
    \item \texttt{iv\_blurred\_intent\_ids}: list the intent IDs whose
          anchors were weakened or removed (typically all of
          $\mathcal{T}$).
\end{itemize}

\vspace{0.5mm}
\textbf{\#\# Step 4: Synthesise the Fully-Vague variant.} \\
\textbf{Goal.} Hide all intent-layer anchors, remove the detail-layer
anchors, and additionally strip a subset of detail units, retaining only
a core-topic phrase of typically 5 to 10 tokens.

\textbf{Operations.}
\begin{itemize}[leftmargin=*, itemsep=2pt, parsep=0pt]
    \item \textit{Predicate anchors} $\rightarrow$ \emph{Generalise or
          delete}.
    \item \textit{Speech-act anchors} $\rightarrow$ \emph{Weaken or
          remove}.
    \item \textit{Output-type anchors} $\rightarrow$ \emph{Delete}.
          For example: ``five bullet points'' $\rightarrow$ $\varnothing$;
          ``give me a report'' $\rightarrow$ drop ``report''.
    \item \textit{Scope anchors} $\rightarrow$ \emph{Hedge or delete}.
          For example: ``analyse from a nutritional perspective''
          $\rightarrow$ ``about nutrition''.
\end{itemize}

\textbf{Rewriting rules.}
\begin{itemize}[leftmargin=*, itemsep=2pt, parsep=0pt]
    \item Identify the most core entity or keyword of the seed and
          express it in a minimalist way, such as
          ``[Core Entity]'' or ``[Core Entity] + short vague modifier''.
    \item Retaining one to two core detail words is allowed, but the
          intent must remain hidden.
    \item The result should read like a casual, partially-typed user
          input.
    \item \texttt{fv\_retained\_detail\_ids}: list the detail IDs that
          remain visible in the Fully-Vague variant; the list may be
          empty.
\end{itemize}

\vspace{0.5mm}
\textbf{\#\# Output constraints.}
\begin{itemize}[leftmargin=*, itemsep=2pt, parsep=0pt]
    \item All id fields, including \texttt{intent\_ids},
          \texttt{iv\_blurred\_intent\_ids}, and
          \texttt{fv\_retained\_detail\_ids}, must reference only the
          ids defined in this same output.
    \item Do not describe any set operations in prose; only emit the
          structured fields.
    \item Do not reason about what was hidden; only record what was
          retained.
\end{itemize}

\textbf{\#\# Output format (strict JSON).}

\vspace{0.5mm}
{\ttfamily
\noindent \{ \\
\hspace*{4mm}"query\_type": "Type", \\
\hspace*{4mm}"intents": [ \\
\hspace*{8mm}\{ \\
\hspace*{12mm}"id": "I1", \\
\hspace*{12mm}"label": "Description of task intent", \\
\hspace*{12mm}"triggers": ["trigger phrase 1", "trigger phrase 2"], \\
\hspace*{12mm}"response": "Colloquial short answer" \\
\hspace*{8mm}\} \\
\hspace*{4mm}], \\
\hspace*{4mm}"details": [ \\
\hspace*{8mm}\{ \\
\hspace*{12mm}"id": "d1", \\
\hspace*{12mm}"key": "Field name", \\
\hspace*{12mm}"value": "Field value", \\
\hspace*{12mm}"intent\_ids": ["I1"], \\
\hspace*{12mm}"triggers": ["trigger word 1", "trigger word 2"], \\
\hspace*{12mm}"response": "Colloquial short answer" \\
\hspace*{8mm}\} \\
\hspace*{4mm}], \\
\hspace*{4mm}"iv\_query": "Intent-Vague variant (intent blurred, all details kept)", \\
\hspace*{4mm}"iv\_blurred\_intent\_ids": ["I1"], \\
\hspace*{4mm}"fv\_query": "Fully-Vague variant (5--10 token core-topic phrase)", \\
\hspace*{4mm}"fv\_retained\_detail\_ids": [] \\
\}
}

\vspace{0.5mm}
\hrule
\vspace{0.5mm}

\textbf{[User Prompt]} \\
The seed query is: \texttt{\{seed\_query\}}
\end{tcolorbox}
\caption{\textbf{Prompt 2: Variant Synthesis.}}
\label{fig:prompt2}
\end{figure*}

\begin{figure*}[!t]
\begin{tcolorbox}[
    colback=gray!5!white,
    colframe=gray!75!black,
]
\textbf{[System Prompt]} \\
Classify the assistant's response to a query as either
\textbf{ask} (a clarifying question) or \textbf{answer} (a direct
attempt at the task). This classification supplies the
\emph{decision-appropriateness} signal that produces the Ask\% column of
Table~\ref{tab:main}.

\begin{itemize}[leftmargin=*, itemsep=2pt, parsep=0pt]
    \item \textbf{ask}: The core purpose of the response is to elicit
          additional information from the user. It contains
          interrogative sentences or explicit information requests
          directed at the user and does not simultaneously provide a
          complete answer to the task.
    \item \textbf{answer}: The response provides a substantive answer
          to the task. Even if an optional follow-up question is
          appended at the end, such as ``these are the recommendations,
          do you have any other preferences?'', the response is still
          considered a direct answer.
\end{itemize}

Output JSON: \texttt{\{"behavior": "ask" | "answer"\}}

\vspace{2mm}
\hrule
\vspace{2mm}

\textbf{[User Prompt]} \\
The user's query is: \texttt{\{query\}} \\
The assistant's response is: \texttt{\{response\}}
\end{tcolorbox}
\caption{\textbf{Prompt 3: Behavior Judge.}}
\label{fig:prompt3}
\end{figure*}

\begin{figure*}[!t]
\begin{tcolorbox}[
    colback=gray!5!white,
    colframe=gray!75!black,
]
\textbf{[System Prompt]} \\
You are an information-extraction expert. You are given a piece of
\emph{text}, the user's \emph{original query}, and a structured
\emph{ground-truth information set} containing an intent list and a
detail list. Your task is to decide, for every intent and every detail
in the ground truth, whether the text covers it. Also list any intent
or detail that the text mentions but the ground truth does not contain.

\textbf{\#\# Rules}
\begin{itemize}[leftmargin=*, itemsep=2pt, parsep=0pt]
    \item Mark an intent or detail as covered only when the text gives
          explicit textual evidence. Do not over-infer.
    \item If the text touches on a topic generally without specifying
          parameters, it still counts as covering the corresponding
          intent.
    \item If the text references content related to a detail with
          different surface wording, the detail still counts as
          covered.
    \item Any intent or detail that appears in the text but not in the
          ground-truth list must be reported under
          \texttt{extra\_intents} or \texttt{extra\_details}.
    \item Return only the JSON object shown below. Do not output any
          explanation or commentary outside the JSON.
\end{itemize}

\textbf{\#\# Output format (strict JSON).}

\vspace{0.5mm}
{\ttfamily
\noindent \{ \\
\hspace*{4mm}"covered\_intent\_ids": ["I1"], \\
\hspace*{4mm}"covered\_detail\_ids": ["d1", "d3"], \\
\hspace*{4mm}"extra\_intents": [ \\
\hspace*{8mm}"Description of intents in the response that are not in
the ground truth" \\
\hspace*{4mm}], \\
\hspace*{4mm}"extra\_details": [ \\
\hspace*{8mm}"Description of details in the response that are not in
the ground truth" \\
\hspace*{4mm}] \\
\}
}

\vspace{2mm}
\hrule
\vspace{2mm}

\textbf{[User Prompt]} \\
The user's original query is: \texttt{\{query\}} \\
The text to evaluate is: \texttt{\{response\}} \\
Ground-truth intent list: \texttt{\{intents\_json\}} \\
Ground-truth detail list: \texttt{\{details\_json\}}
\end{tcolorbox}
\caption{\textbf{Prompt 4: Information Extractor.}}
\label{fig:prompt4}
\end{figure*}

\end{document}